%% file: arxiv.tex
\documentclass{article} 

\usepackage{amsmath,amssymb,amsthm,amsfonts}
\usepackage{caption,subcaption}
\usepackage{times}
\usepackage{xr}
\usepackage{hyperref}
\usepackage{url}
\usepackage{amssymb}
\usepackage[round]{natbib}
\usepackage{amsbsy}
\usepackage{amsthm}
\usepackage{amsmath}
\usepackage{paralist}
\usepackage{bm}
\usepackage{booktabs}
\usepackage{rotating}
\usepackage{cases}
\usepackage{mathtools}
\usepackage{graphicx}
\usepackage{setspace}
\usepackage{textcomp}
\usepackage[accepted]{arxivicml2016}

\newcommand{\R}{\mathbb{R}}
\newcommand{\vnorm}[1]{\left|\left|#1\right|\right|}

\icmltitlerunning{GapTV: Accurate and Interpretable Low-Dimensional Regression and Classification}

\begin{document}

\twocolumn[
\icmltitle{GapTV: Accurate and Interpretable Low-Dimensional Regression and Classification}

\icmlauthor{Wesley Tansey}{tansey@cs.utexas.edu}
\icmladdress{Department of Computer Science,
            University of Texas at Austin}
\icmlauthor{James G.~Scott}{james.scott@mccombs.utexas.edu}
\icmladdress{Department of Information, Risk, and Operations Management; 
            Department of Statistics and Data Sciences,
            University of Texas at Austin}

\icmlkeywords{interpretable regression, convex clustering, total variation denoising, graph fused lasso}

\vskip 0.3in
]

\begin{abstract}
We consider the problem of estimating a regression function in the common situation where the number of features is small, where interpretability of the model is a high priority, and where simple linear or additive models fail to provide adequate performance.  To address this problem, we present GapTV, an approach that is conceptually related both to CART and to the more recent CRISP algorithm \citep{petersen:etal:2016}, a state-of-the-art alternative method for interpretable nonlinear regression.  GapTV divides the feature space into blocks of constant value and fits the value of all blocks jointly via a convex optimization routine.  Our method is fully data-adaptive, in that it incorporates highly robust routines for tuning all hyperparameters automatically.  We compare our approach against CART and CRISP and demonstrate that GapTV finds a much better trade-off between accuracy and interpretability.
\end{abstract}

\input{introduction}

\input{background}

\input{algorithm}

\input{experiments}

\input{conclusion}

\begin{small}
\bibliographystyle{icml2016}
\bibliography{crispier}
\end{small}

\appendix

\title{GapTV - Appendix}
\date{} 

\maketitle

\input{app_chicago}

\end{document}

%% file: introduction.tex
\section{Introduction}
\label{sec:introduction}
Many modern machine learning techniques, such as deep learning and kernel machines, tend to focus on the ``big data, big features'' regime. In such a scenario, there are often so many features and highly non-linear interations between features that model interpretability is generally a secondary consideration. Instead, effort is focused soley on a measure of model performance such as root mean squared error (RMSE). Under this research paradigm, only a model that out-performs the previous champion method warrants an investigation into understanding its decisions.

But there is also a robust and recent line of machine-learning research in the equally important scenario of low-dimensional regression, with relatively few features and where interpretability is a primary concern. For example, lattice regression with monotonicity constraints has been shown to perform well in video-ranking tasks where interpretability was a prerequisite \cite{gupta:etal:2016}. The interpretability of the system enables users to investigate the model, gain confidence in its recommendations, and guide future recommendations. In the two- and three- dimensional regression scenario, the Convex Regression via Interpretable Sharp Partitions (CRISP) method \cite{petersen:etal:2016} has recently been introduced as a way to achieve a good trade off between accuracy and interpretability by inferring sharply-defined 2d rectangular regions of constant value. Such a method is readily useful, for example, when making business decisions or executive actions that must be explained to a non-technical audience.  CRISP is similar to classification and regression trees (CART), in that it partitions the feature space into contiguous blocks of constant value (``interpretable sharp partitions''), but was shown to lead to better performance.

Another area where data-adaptive, interpretable sharp partitions are useful is in the creation of areal data from a set of spatial point-referenced data---essentially turning a continuous spatial problem into a discrete one.  A common application of the framework arises when dividing a city, state, or other region into a set of contiguous cells, where values in each cell are aggregated to help anonymize individual demographic data.  Ensuring that the number and size of grid cells remains tractable, handling low-data regions, and preserving spatial structure are all important considerations for this problem. Ideally, one cell should contain data points which all map to a similar underlying value, and cell boundaries should represent significant change points in the value of the signal being estimated. If a cell is empty or contains a small number of data points, the statistical strength of its neighbors should be leveraged to both improve the accuracy of the reported areal data and further aide in anonymizing the cell which may otherwise be particularly vulnerable to deanonymization.  Viewed through this lens, we can interpret the areal-data creation task as a machine learning problem, one focused on finding sharp partitions that still achieve acceptable predictive loss.\footnote{We note that such a task will likely only represent a single step in a larger anonymization pipeline that may include other techniques such as additive noise and spatial blurring. While we provide no proofs of how strong the anonymization is for our method, we believe it is compatible with other methods that focus on adherence to a specified \textit{k}-anonymity threshold (e.g., \cite{cassa:etal:2006}).}

To this end, and motivated by the success of CRISP, we present GapTV, a method for interpretable, low-dimensional convex regression with sharp partitions. GapTV involves two main steps: (1) a non-standard application of the gap statistic \cite{tibshirani:etal:2001} to create a data-adaptive grid over the feature space; and (2) smoothing over this grid using a fast total variation denoising algorithm \cite{barbero:sra:2014}. The resulting model displays a good balance between four key measurements: (1) interpretability, (2) average accuracy, (3) worst-region accuracy, and (4) degrees of freedom. Through a series of benchmarks against both a baseline CART model and the state-of-the-art CRISP model, we show both qualitatively and quantitatively that GapTV achieves superior performance. The end result is a fast, fully auto-tuned approach to interpretable low-dimensional regression and classification.

The remainder of this paper is organized as follows. Section \ref{sec:background} presents technical background on both CRISP and graph-based total variation denoising. In Section \ref{sec:algorithm}, we detail our algorithm and derive the gap statistic for both regression and classification scenarios. We then present a suite of benchmark experiments in Section \ref{sec:experiments} and conclude in Section \ref{sec:conclusion}.

%% file: background.tex
\section{Background}
\label{sec:background}


\subsection{Convex Regression with Interpretable Sharp Partitions}
\label{subsec:crisp}

\citet{petersen:etal:2016} propose the CRISP algorithm for handling the prediction scenario described previously. As in our approach, they focus on the 2d scenario and divide the $(x_1, x_2)$ space into a grid via a data-adaptive procedure. For each dimension, they divide the space into $q$ regions, where each region break is chosen such that a region contains $1/q$ of the data. This creates a $q \times q$ grid of differently-sized cells, some of which may not contain any observations. A prediction matrix $M \in \R^{q \times q}$ is then learned, with each element $M_{ij}$ representing the prediction for all observations in the region specified by cell $(i,j)$.

CRISP applies a Euclidean penalty on the differences between adjacent rows and columns of $M$. The final estimator is then learned by solving the convex optimization problem
\begin{equation}
\label{eqn:crisp_objective}
\begin{aligned}
& \underset{M \in \R^{q \times q}}{\text{minimize}}
& & 
\frac{1}{2}\sum_{i = 1}^n (y_i - \Omega(M, x_{1i}, x_{2i}))^2 + \lambda P(M) \, ,
\end{aligned}
\end{equation}
where $\Omega$ is a lookup function mapping $(x_{1i}, x_{2i})$ to the corresponding element in $M$. $P(M)$ is the group-fused lasso penalty on the rows and columns of $M$
\begin{equation}
\label{eqn:crisp_penalty}
P(M) = \sum_{i = 1}^{q-1} \left[ \vnorm{M_{i\cdot} - M_{(i+1)\cdot}}_2 + \vnorm{M_{\cdot i} - M_{\cdot(i+1)}}_2 \right] \, ,
\end{equation}
where $M_{i\cdot}$ and $M_{\cdot i}$ are the $i^{\text{th}}$ row and column of $M$, respectively.

By rewriting $\Omega(\cdot)$ as a sparse binary selector matrix and introducting slack variables for each row and column in the $P(M)$ term, CRISP solves \eqref{eqn:crisp_objective} via ADMM. The resulting algorithm requires an initial step of $\mathcal{O}(n+q^4)$ operations for $n$ samples on a $q\times q$ grid, and has a per-iteration complexity of $\mathcal{O}(q^3)$. The authors recommend using $q=n$ when the size of the data is sufficiently small so as to be computationally tractable, and setting $q=100$ otherwise.

In comparison to other interpretable methods, such as CART and thin-plate splines (TPS), CRISP is shown to yield a good tradeoff between accuracy and interpretability. Consequently, we use CRISP as our main method to compare against in Section \ref{sec:experiments}.

\subsection{Graph-based Total Variation Denoising}
\label{subsec:graphtv}

Total variation (TV) denoising solves a convex regularized optimization problem defined generally over a graph $\mathcal{G} = (\mathcal{V}, \mathcal{E})$ with node set $\mathcal{V}$ and edge set $\mathcal{E}$:
\begin{equation}
\label{eqn:gfl_objective}
\begin{aligned}
& \underset{\boldsymbol\beta \in \R^{|\mathcal{V}|}}{\text{minimize}}
& & 
\sum_{s \in \mathcal{V}} \ell(\beta_s) + \lambda \sum_{(r,s) \in \mathcal{E}} |\beta_r - \beta_s| \, ,
\end{aligned}
\end{equation}
where $\ell$ is some smooth convex loss function over the value a given node $\beta_s$. The solution to \eqref{eqn:gfl_objective} yields connected subgraphs (i.e. plateaus in the 2d case) of constant value. TV denoising has been shown to have attractive minimax rates theoretically \cite{wang:etal:2014} and is robust against model mispecification empirically, particularly in terms of worst-cell error \cite{tansey:scott:2016:multiscale}.

Many efficient, specialized algorithms have been developed for the case when $\ell$ is a Gaussian loss and the graph has a specific constrained form. For example, when $\mathcal{G}$ is a one-dimensional chain graph, \eqref{eqn:gfl_objective} is the ordinary (1D) fused lasso \cite{tibs:fusedlasso:2005}, solvable in linear time via dynamic programming \cite{johnson:2013}. When $\mathcal{G}$ is a D-dimensional grid graph, \eqref{eqn:gfl_objective} is typically referred to as total variation denoising \cite{rudin:osher:faterni:1992} or the graph-fused lasso, for which several efficient solutions have been proposed \cite{chambolle:darbon:2009,barbero:sra:2011,barbero:sra:2014}. For scenarios with a general smooth convex loss and an arbitrary graph, the GFL method \cite{tansey:scott:2015gfl} is efficient and easily extended to non-Gaussian losses such as the binomial loss required in Section \ref{subsec:algorithm_classification}.

The TV denoising penalty was investigated as an alternative to CRISP in \cite{petersen:etal:2016}. They note anecdotally that TV denoising over-smooths when the same $q$ was used for both CRISP and TV denoising. In the next section, we present a principled approach to choosing $q$ in a data-adaptive way that prevents over-smoothing and leads to a superior fit in terms of the accuracy-interpretability tradeoff.

%% file: algorithm.tex
\section{The GapTV Algorithm}
\label{sec:algorithm}

Prior to presenting our approach, we first note that we can rewrite \eqref{eqn:crisp_objective} as a weighted least-squares problem
\vspace{-0.1in}\begin{equation}
\label{eqn:weighted_least_squares_objective}
\begin{aligned}
& \underset{\boldsymbol\beta \in \R^{q^2}}{\text{minimize}}
& & 
\frac{1}{2}\sum_{i = 1}^{q^2} \eta_i (\tilde{y}_i - \beta_i)^2 + \lambda g(\boldsymbol\beta) \, ,
\end{aligned}
\vspace{-0.1in}
\end{equation}
where $\boldsymbol\beta = \text{vec}(M)$ is the vectorized form of $M$, $\eta_i$ is the number of observations in the $i^{\text{th}}$ cell, and $\tilde{y}_i$ is the empirical average of the observations in the $i^{\text{th}}$ cell. $g(\cdot)$ is then a penalty term that operates over a vector $\boldsymbol\beta$ rather than a matrix $M$.

Given the reformulation of the problem in \eqref{eqn:weighted_least_squares_objective}, we now choose $g(\cdot)$ to be a graph-based total variation penalty
\vspace{-0.1in}\begin{equation}
\label{eqn:graph_tv_penalty}
g(\boldsymbol\beta) = \sum_{(r,s) \in \mathcal{E}} |\beta_r - \beta_s| \, ,
\vspace{-0.1in}\end{equation}
where $\mathcal{E}$ is the set of edges defining adjacent cells on the $q \times q$ grid graph.\footnote{Though our goal in this work is not to increase the computational efficiency of existing methods, we do note that CRISP can be solved substantially faster via the reformulation in \eqref{eqn:weighted_least_squares_objective}. The weighted least squares loss enables a much more efficient solution to \eqref{eqn:crisp_objective} via a simpler ADMM solution similar to the network lasso \cite{hallac:etal:2015}.} Having formulated the problem as a graph TV denoising problem, we can now use the convex minimization algorithm of \citet{barbero:sra:2014} (or any other suitable algorithm) to efficiently solve \eqref{eqn:weighted_least_squares_objective}.

The remainder of this section is dedicated to our approach to auto-tuning the two hyperparameters: $q$, the granularity of the grid, and $\lambda$, the regularization parameter. We take a pipelined approach by first choosing $q$ and then selecting $\lambda$ under the chosen $q$ value.

\subsection{Choosing bins via the gap statistic}
\label{subsec:choosing_q}
The recommendation for CRISP is to choose $q = n$, assuming the computation required is feasible. Doing so creates a very sparse grid, with $q-1 \times q$ empty cells. However, by tying together the rows and columns of the grid, each CRISP cell actually draws statistical strength from a large number of bins. This compensates for the data sparsity problem and results in reasonably good fits despite the sparse grid.

Unfortunately, choosing $q = n$ does not work for our TV denoising approach. Since the graph-based TV penalty only ties together adjacent cells, long patches of sparsity overwhelm the model and result in over-smoothing. If one instead chooses a smaller value of $q$, however, the TV penalty performs quite well. The challenge is therefore to adaptively choose $q$ to fit the appropriate level of overall data sparsity. We propose to do this via a novel use of the gap statistic \citep{tibshirani:etal:2001}.

In a typical clustering algorithm, such as $K$-means, one would have unlabeled data $X = \{\mathbf{x}_1, \mathbf{x}_2, \ldots, \mathbf{x}_n\}$, some distance metric $\delta(\mathbf{x}_i, \mathbf{x}_j)$, and a specified number of $K$ clusters to find. In $K$-means, cluster assignment is based on the nearest centroid,
\begin{equation}
\label{eqn:kmeans_assignment}
a_i = \underset{k}{\text{argmin}}\quad \delta(\mathbf{x}_i, \mathbf{c}_k) \, ,
\end{equation}
where $\mathbf{c}_k = \frac{1}{|A_k|} \sum_{i \in A_k} \mathbf{x}_i$ is the cluster centroid and $A_k = \{i : a_i = k, \forall i\}$.

The gap statistic is an approach to choosing the value of $K$ for a generic clustering algorithm by comparing it against a suitable null distribution. The best clustering is the one which minimizes the gap term:
\begin{equation}
\label{eqn:gap_statistic}
\mathbb{E}_n\left[\log(W^*_1)\right] - log(W_K) \, ,
\end{equation}
where $W_K$ is the sum of average pairwise distances in each cluster for a clustering with $K$ clusters. To use the gap statistic, one must define a suitable null distribution over $W_1$.

In our case, the ``clusters'' are defined by a quantile grid over $(x_1, x_2)$. The number of cells is specified by the choice of $q$, which means choosing the value of $q$ corresponds directly to choosing $K$. However, unlike typical clustering, a cluster centroid is defined by the $y_i$ values corresponding to the $\mathbf{x}_i$ points in the cell. Therefore, our distance metric for computing the gap statistic is actually between pairs of $(y_i, y_j)$.

In the regression case, we assume each $y_i \sim \mathcal{N}(\mu, \sigma^2)$, where $\mu$ and $\sigma^2$ are unknown. For a distance metric, we use Euclidean distance,
\begin{equation}
\label{eqn:gap_distance}
\delta(y_i, y_j) = (y_i - y_j)^2 \, .
\end{equation}
Since each $y_i$ is assumed to be IID normal, the null distribution over pairwise distances is $W_1 \sim 2\sigma^2 \chi^2_\nu$, where $\nu = \frac{n^2}{2} - n$ is the degrees of freedom. The expectation of the log of a $\chi^2$ distribution can be calculated exactly \citep{walck:2007} as
\begin{equation}
\label{eqn:expected_log_chi_sq}
\mathbb{E}\left[\log(\chi^2_\nu)\right] = \log 2 + \psi\left(\frac{\nu}{2}\right) \, ,
\end{equation}
where $\psi$ is the digamma function. Thus, up to an additive constant, we can calculate the reference distribution exactly without knowing the mean or variance.

The procedure for choosing $q$ is now straightforward. We first partition the points on a grid for a series of candidate $q$ values in the range $1 < q \leq q_{\text{max}} \leq n$. For each candidate partitioning, we calculate the gap statistic
\begin{equation}
\label{eqn:crisp_gap_statistic}
\text{gap}(q) = \psi(\frac{\nu}{2}) - \sum_{k = 1}^{q^2}\frac{1}{\eta_k}\sum_{i \in A_i}\sum_{j \in A_i,\\j>i} \delta(y_i, y_j) \, .
\end{equation}
We then choose the $q$ which minimizes $\text{gap}(q)$ and smooth using the TV denoising algorithm.

\subsection{Choosing the TV penalty parameter}
\label{subsec:choosing_lambda}

Once a value of $q$ has been chosen, $\lambda$ can be chosen by following a solution path approach. For the regression scenario with a Gaussian loss, as in \eqref{eqn:weighted_least_squares_objective}, determining the degrees of freedom is well studied \citep{tibs:taylor:2011}. Thus, we could select $\lambda$ via an information criterion such as AIC or BIC. However, we chose to select $\lambda$ via cross-validation as we found empirically that it produces better results.

\subsection{Classification extension}
\label{subsec:algorithm_classification}

The optimization problem in \eqref{eqn:weighted_least_squares_objective} focuses purely on the Gaussian loss case. When the observations are binary labels, as in classification, a binomial loss function is a more appropriate choice. The binomial loss case specifically has been derived in previous work \citep{tansey:scott:2016:multiscale} and shown to be robust to numerous types of underlying spatial functions. Therefore, unlike CRISP, the inner loop of our method immediately generalizes to the non-Gaussian scenario, with only minor modifications.

In order to adapt the gap statistic to the binomial case, we must find a suitable reference distribution. We assume every $y_i$ is Bernoulli distributed, from which it follows:
\begin{subequations}
\begin{align}
y_i, y_j &\sim \text{Bern}(p) 
\label{eqn:bernoulli_assumption}\\
(y_i - y_j)^2 &\sim \text{Bern}(2p(1-p)) 
\label{eqn:bernoulli_distance}\\
W_1 &\sim \text{Bin}\left(\frac{n^2 - n}{2}, 2p(1-p)\right)
\label{eqn:binomial_reference_distribution} \, .
\end{align}
\end{subequations}
Calculating the expectation of the log of a Binomial in closed form is not tractable, however we can make a close approximation via a Taylor expansion,
\begin{equation}
\label{eqn:expected_logbinomial}
\mathbb{E}\left[\log W_1\right] \approx \log(r*m) - \frac{1-r}{2*r*m} \, ,
\end{equation}
where $m = \frac{n^2 - n}{2}$ and $r = 2p(1-p)$.

Extensions to any other smooth, convex loss are straightforward. One must simply define a loss and a probabilistic model for each data point. Depending on the choice of model, the expectation of the log of the null may not always have a closed form solution. In such cases, we suggest following the simulation strategy specified in \citep{tibshirani:etal:2001}.

%% file: experiments.tex
\section{Experiments}
\label{sec:experiments}

To evaluate the efficacy of our approach, we compare against a suite of both synthetic and real-world datasets. We first compare GapTV against two benchmark methods with sharp partitions, CART and CRISP, on a synthetic dataset with varying sample sizes. We also compare against CRISP with $q$ fixed at the gap statistic solution in a method we call GapCRISP. We show that the GapTV method has much better interpretability qualitatively and leads to better AIC scores. We then demonstrate the advantage of the gap statistic by showing that it chooses grid sizes that offer a good trade-off between average and worst-cell accuracy. Finally, we test all four methods against two real-world datasets of crime reports for Austin and Chicago.

\begin{figure}
\centering
\includegraphics[width=0.3\textwidth]{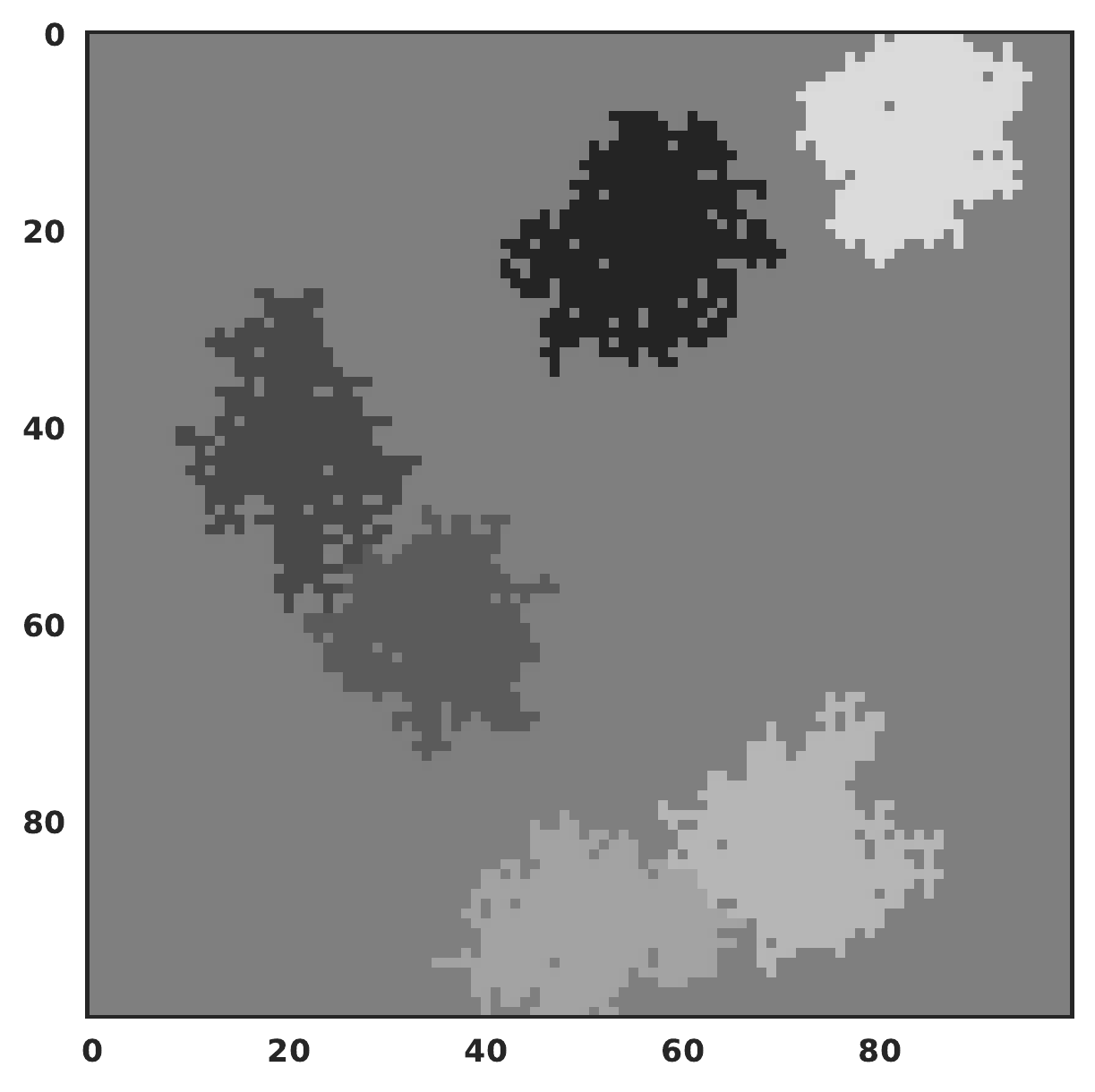}
\vspace{-0.15in}\caption{\label{fig:plateau_example} An example $100\times100$ grid of ground truth means ranging from $-5$ to $5$. Each grid has six randomly-generated plateaus of raised or lowered means from the background mean (zero); darker colors correspond to regions of higher value.\vspace{-0.2in}}
\end{figure} 

\subsection{Synthetic Benchmark}
\label{subsec:experiments:synthetic}
We generated 100 independent $100 \times 100$ grids, each with six 1000-point plateaus. Each plateau was generated via a random walk from a randomly chosen start point and the means of the plateaus were -5, -3, -2, 2, 3, and 5; all points not in a plateau had mean zero. For each grid, we sampled points uniformly at random with replacement and added Gaussian noise with unit variance. Figure \ref{fig:plateau_example} shows an example ground truth for the means. Sample sizes explored for each grid were 50, 100, 200, 500, 1000, 2000, 5000, and 10000. For each trial, we evaluate the CART method from the R package \texttt{rpart}, CRISP, and the Gap* methods. For CRISP, we use $q = \text{max}(n,100)$ as per the suggestions in \cite{petersen:etal:2016}; for the Gap* methods, we use the gap statistic to choose from $q \in [2,50]$. For both CRISP and the Gap* methods, we chose $\lambda$ via 5-fold cross validation across a log-space grid of 50 values.

In order to quantify interpretability, we calculate the number of constant-valued plateaus in each model. Intuitively, this captures the notion of ``sharpness'' of the partitions by penalizing smooth partitions for their visual blurriness. Statistically, this corresponds directly to the degrees of freedom of a TV denoising model in the unweighted Gaussian loss scenario \cite{tibs:taylor:2011}. Thus for all of our models this is only an approximation to the degrees of freedom. Nonetheless, we find the plateau-counting heuristic to be a useful measurement of the \textit{visual} degrees of freedom which corresponds more closely to human interpretability. Finally, to quantify the trade-off of accuracy and interpretability, we use the Akaike information criterion (AIC) with the plateau count as the degrees of freedom surrogate.

\begin{figure*}[!ht]
\centering
\begin{subfigure}[t]{0.32\textwidth}
\includegraphics[width=\textwidth]{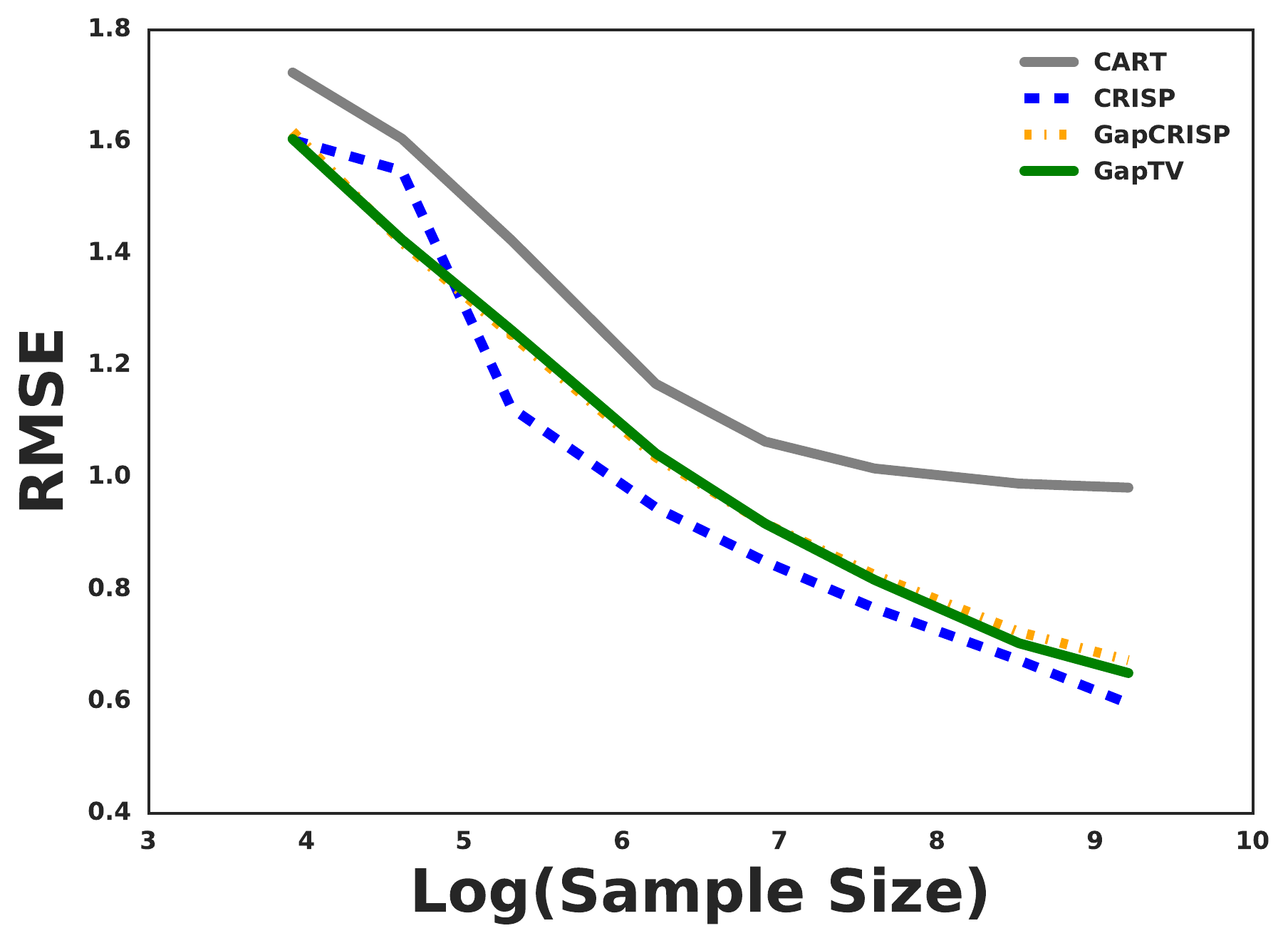}
\caption{RMSE}
\end{subfigure}
\begin{subfigure}[t]{0.32\textwidth}
\includegraphics[width=\textwidth]{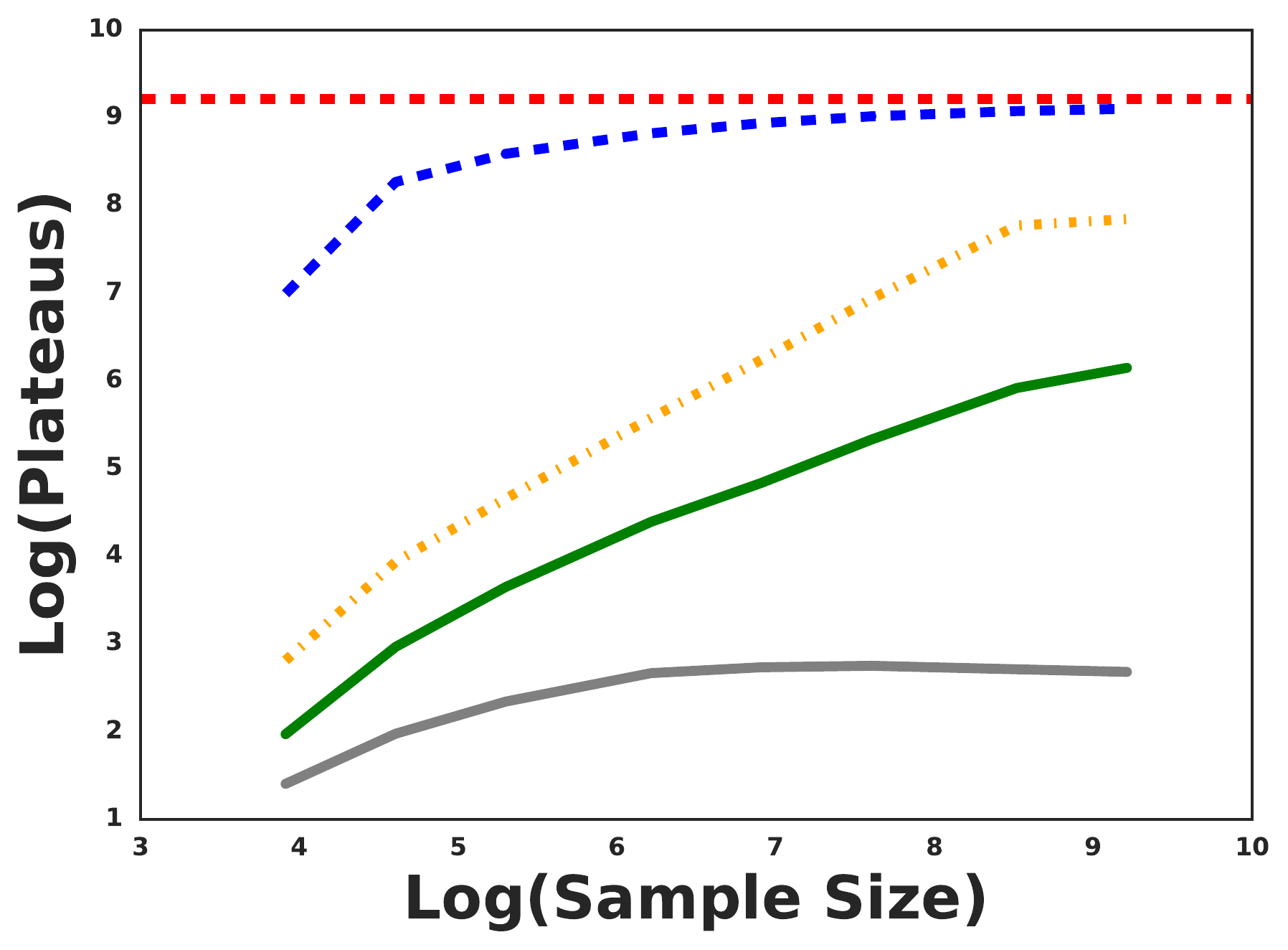}
\caption{Plateau Count}
\end{subfigure}
\begin{subfigure}[t]{0.32\textwidth}
\includegraphics[width=\textwidth]{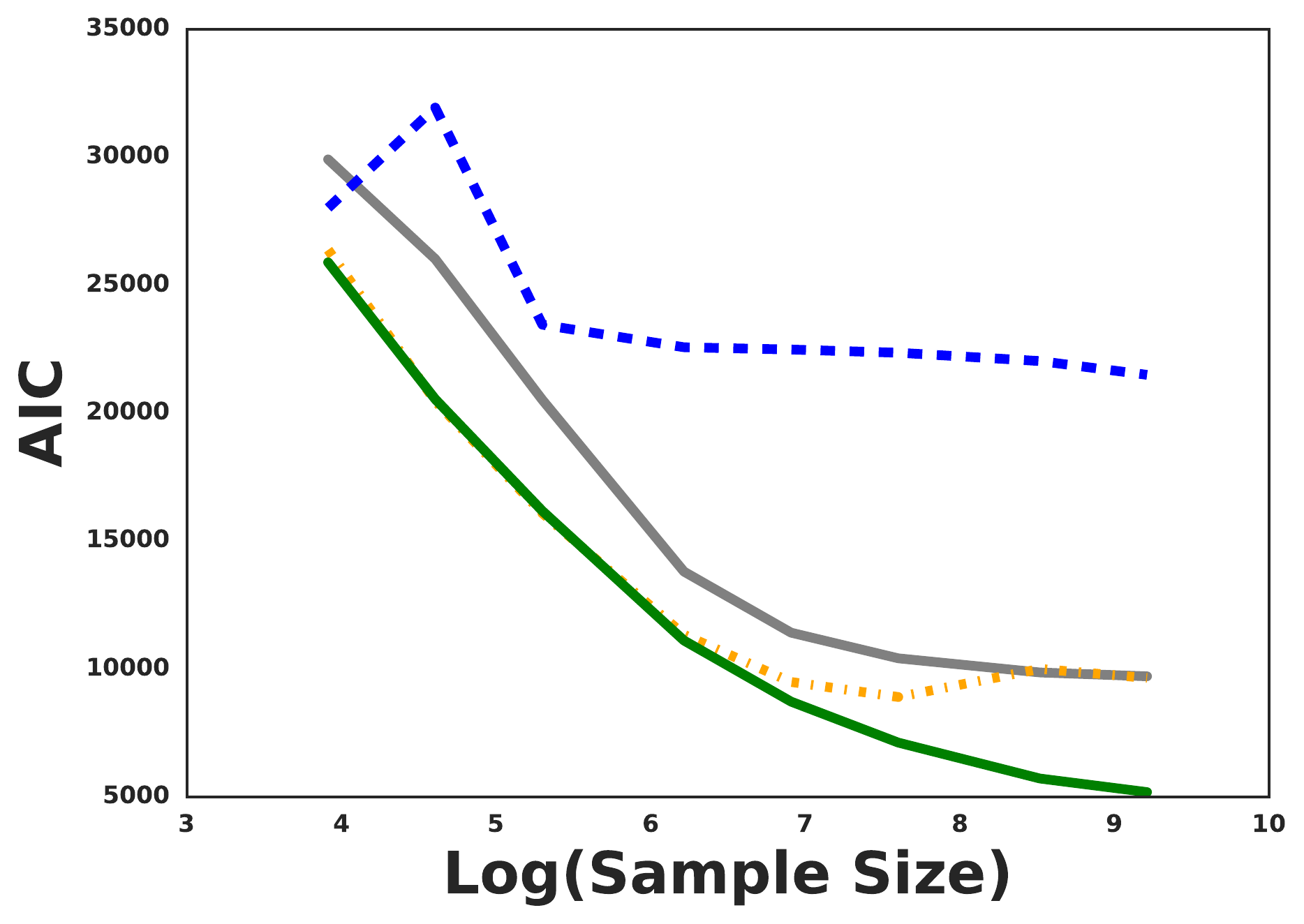}
\caption{Approximate AIC}
\end{subfigure}
\caption{\label{fig:synthetic_results_quantitative} Performance of the four methods as the sample size increases for the example grid in Figure \ref{fig:plateau_example}. While CRISP, GapCRISP, and GapTV achieve similar sample efficiency in terms of RMSE scores (panel A), CRISP and GapCRISP do so with drastically more change points (panel B); the dashed red horizontal line marks the maximum number of plateaus possible. Using AIC as a trade-off measurement (Panel C), both Gap* methods initially perform similarly but as the sample size (and thus the size of $q$) grows, the GapTV method continues to improve while the GapCRISP method begins to over-smooth.}
\end{figure*}

Figure \ref{fig:synthetic_results_quantitative} shows the quantitative results of the experiments, averaged over the 100 trials. The CRISP and Gap* methods perform similarly in terms of RMSE (Figure \ref{fig:synthetic_results_quantitative}a), but both CRISP methods create drastically more plateaus. In the case of the original CRISP method, it quickly approaches one plateau per cell (i.e., completely smooth) as denoted by the dotted red horizontal line in Figure \ref{fig:synthetic_results_quantitative}b. GapTV also presents a better trade-off point as measured by AIC (Figure \ref{fig:synthetic_results_quantitative}c). Using the data-adaptive $q$ value chosen by our gap statistic method helps improve the AIC scores in the low-sample regime, but as samples grow the GapCRISP method begins to under-smooth by creating too many plateaus. This demonstrates that it is not merely the size of the grid, but also our choice of TV-based smoothing that leads to strong results.

\begin{figure*}
\centering
\begin{subfigure}[t]{0.32\textwidth}
\includegraphics[width=\textwidth]{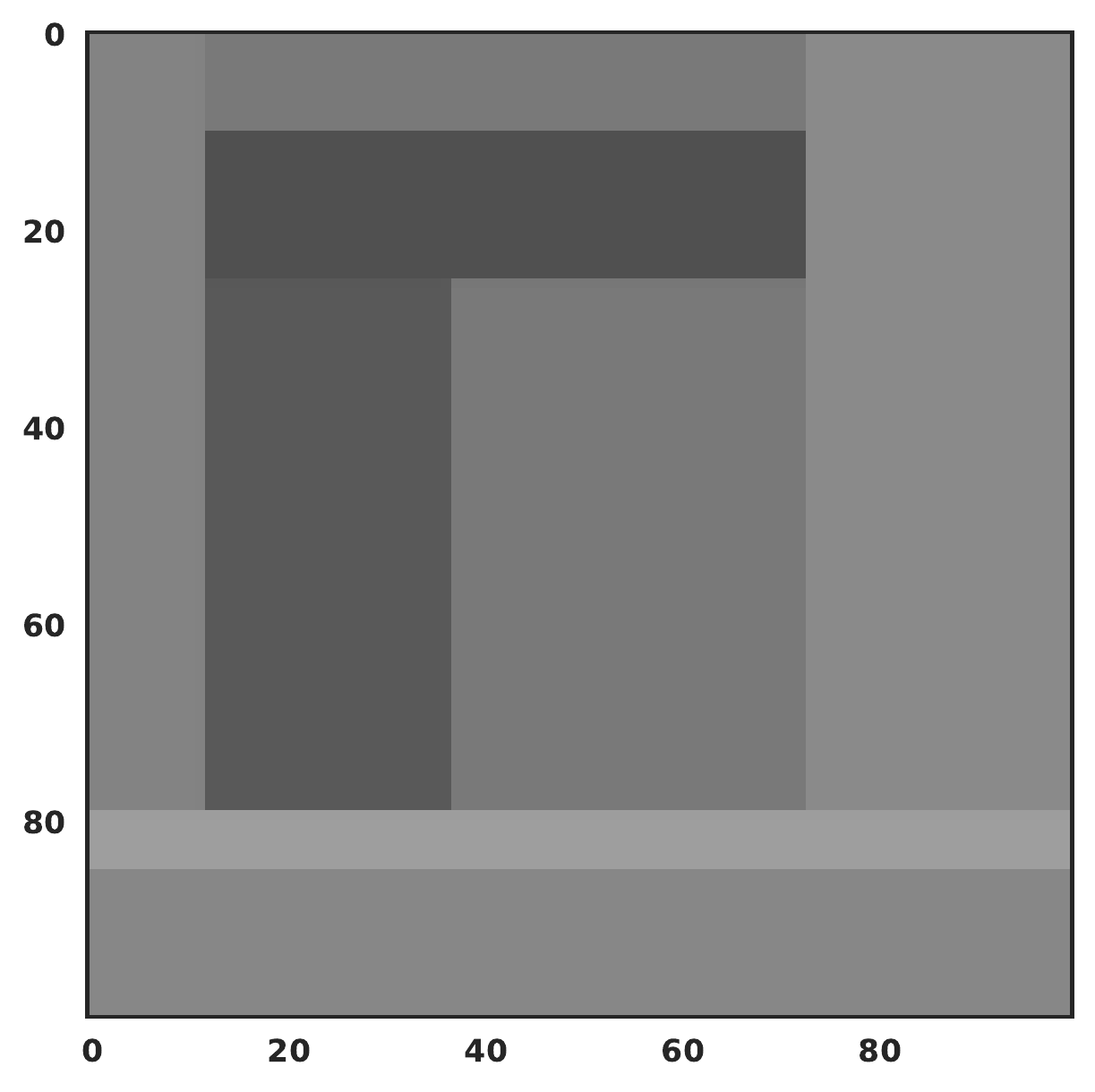}
\caption{CART, N = 100}
\end{subfigure}
\begin{subfigure}[t]{0.32\textwidth}
\includegraphics[width=\textwidth]{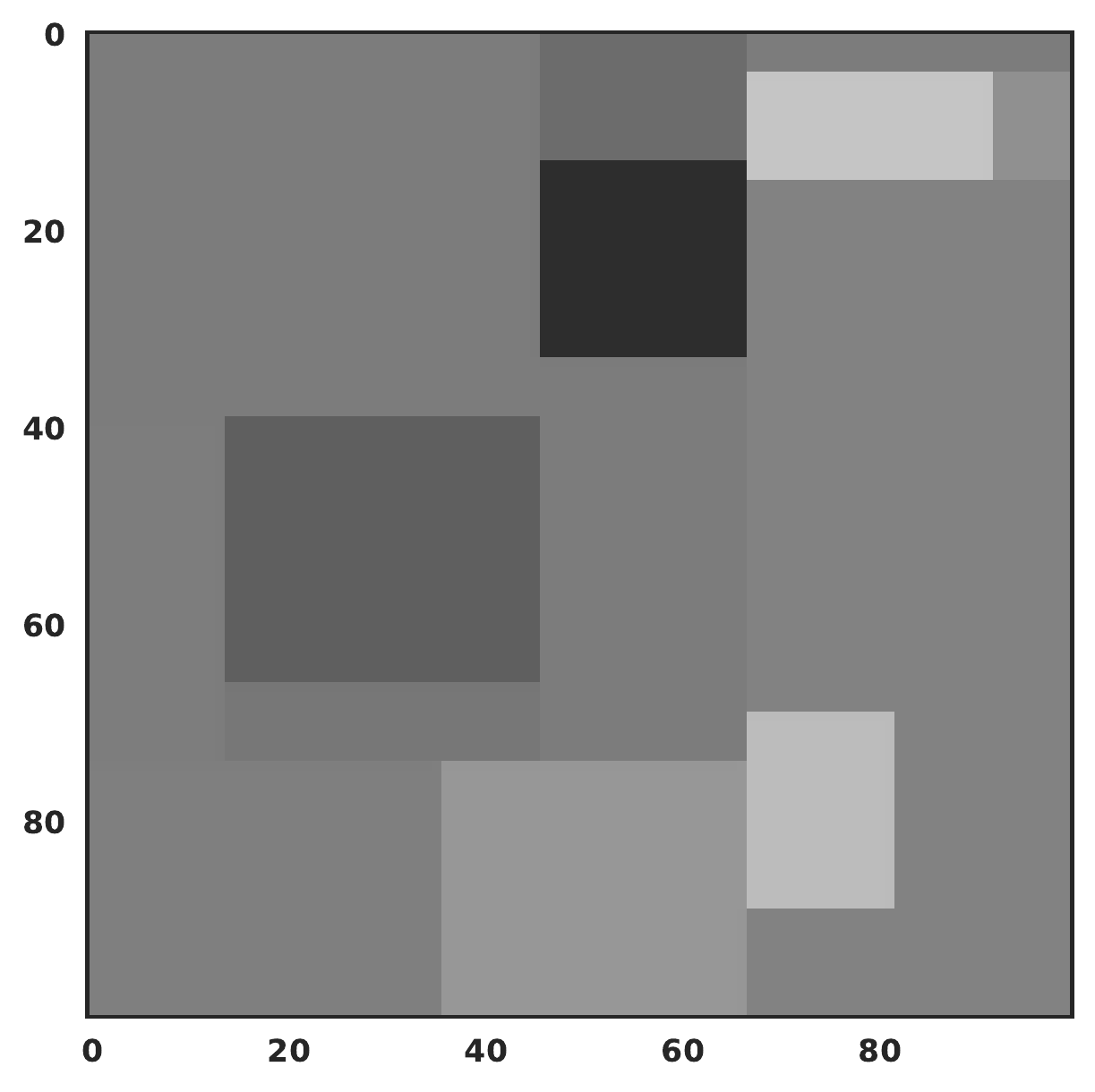}
\caption{CART, N = 500}
\end{subfigure}
\begin{subfigure}[t]{0.32\textwidth}
\includegraphics[width=\textwidth]{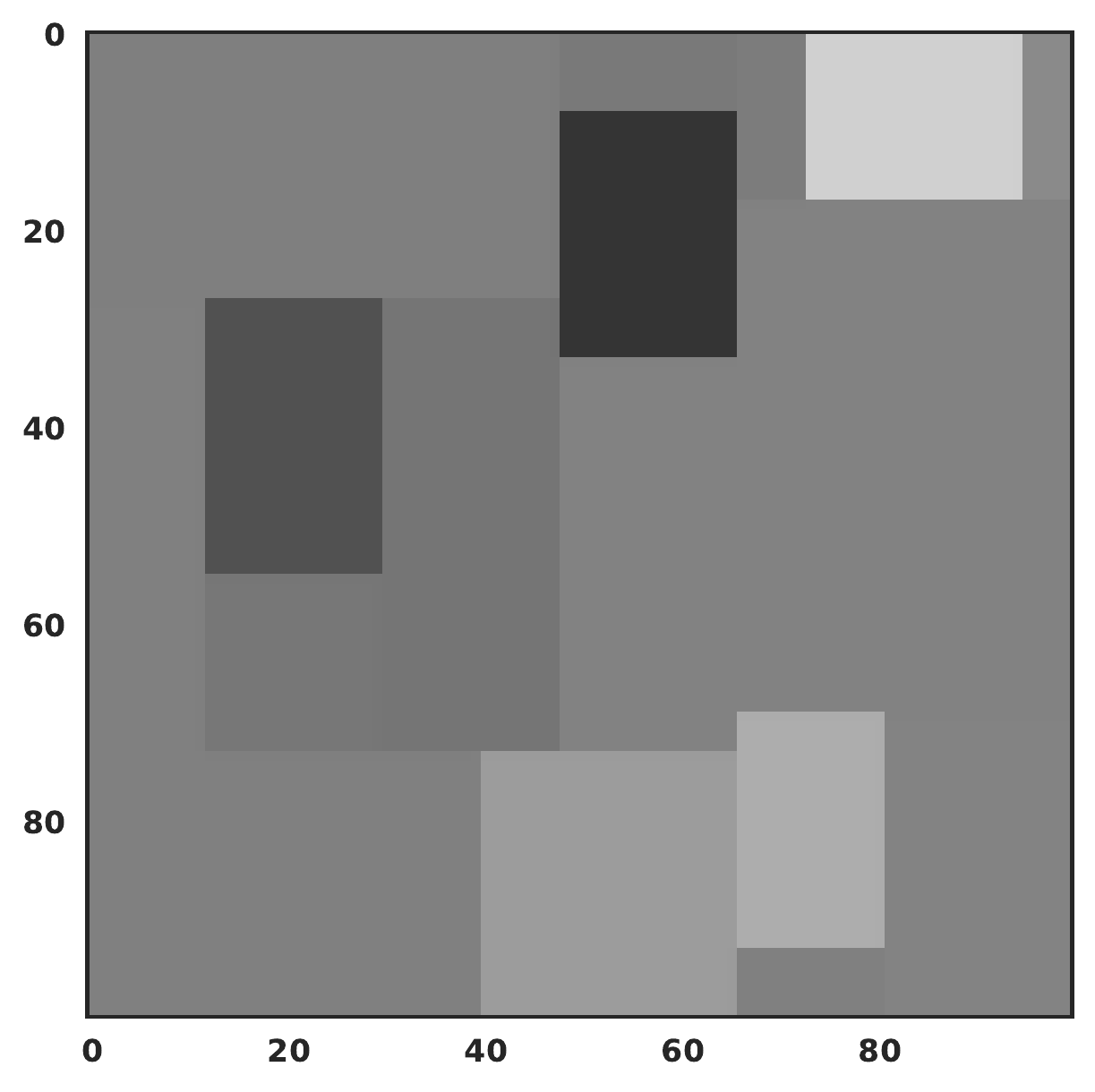}
\caption{CART, N = 2000}
\end{subfigure}
\begin{subfigure}[t]{0.32\textwidth}
\includegraphics[width=\textwidth]{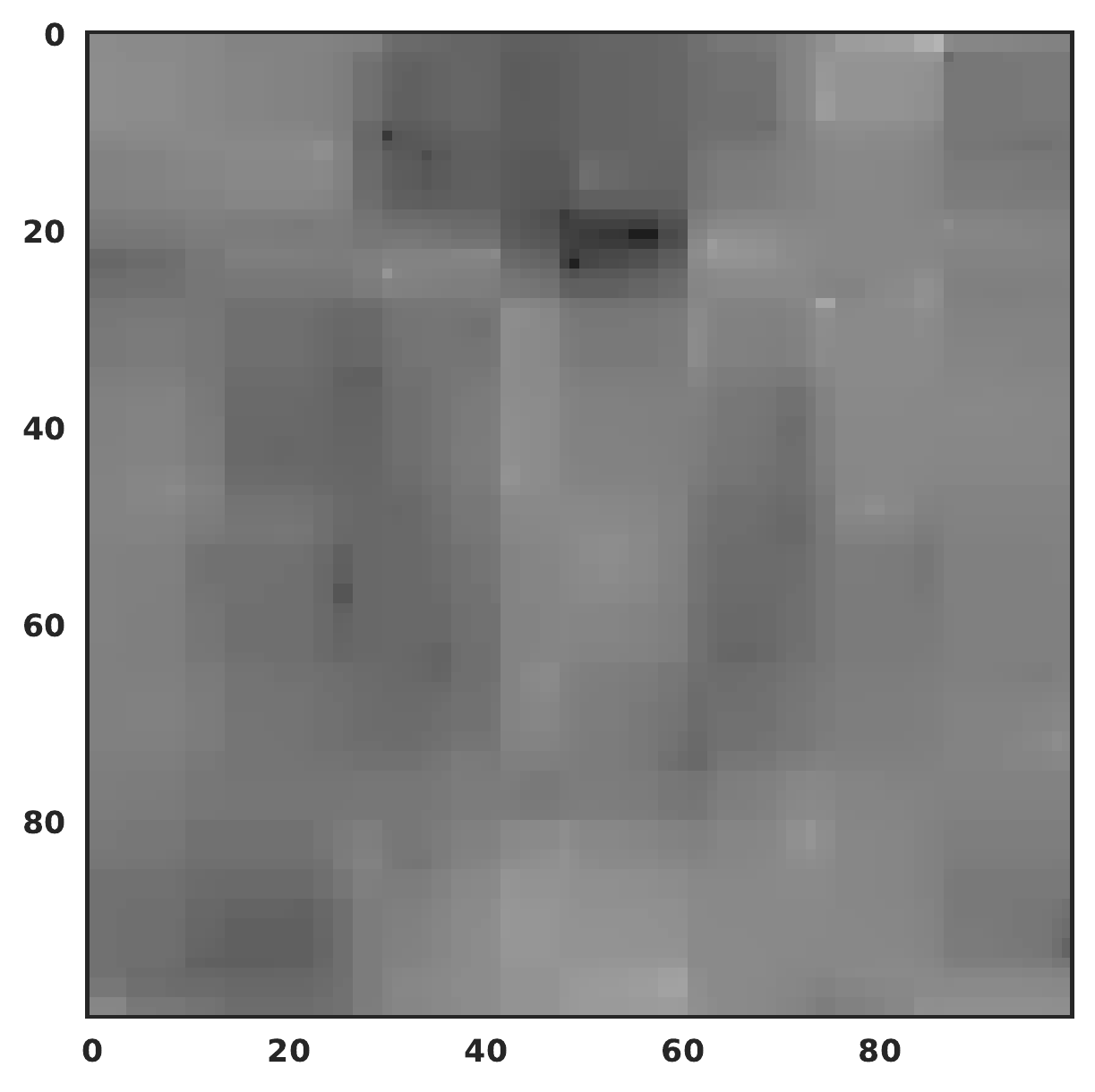}
\caption{CRISP, N = 100}
\end{subfigure}
\begin{subfigure}[t]{0.32\textwidth}
\includegraphics[width=\textwidth]{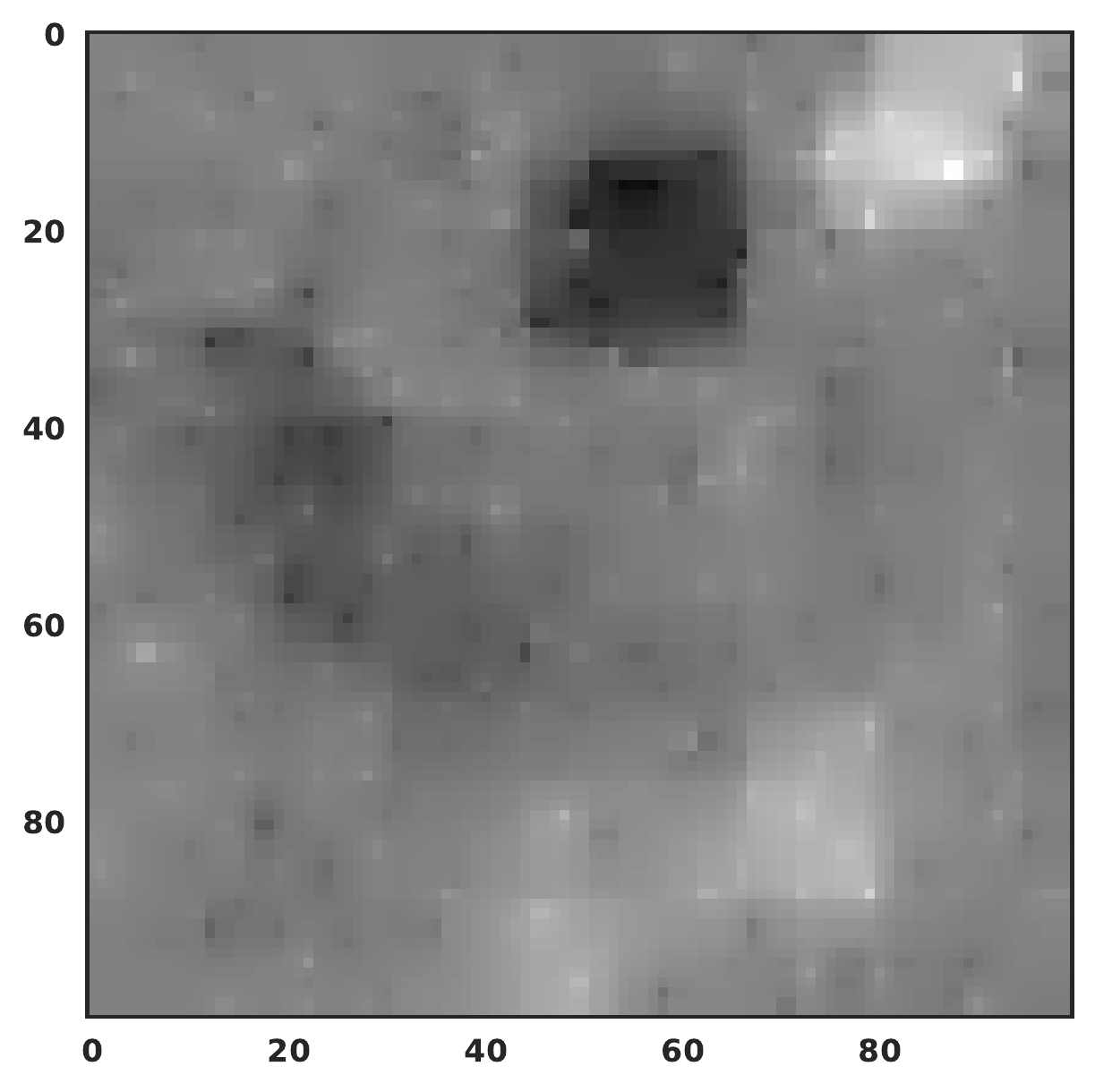}
\caption{CRISP, N = 500}
\end{subfigure}
\begin{subfigure}[t]{0.32\textwidth}
\includegraphics[width=\textwidth]{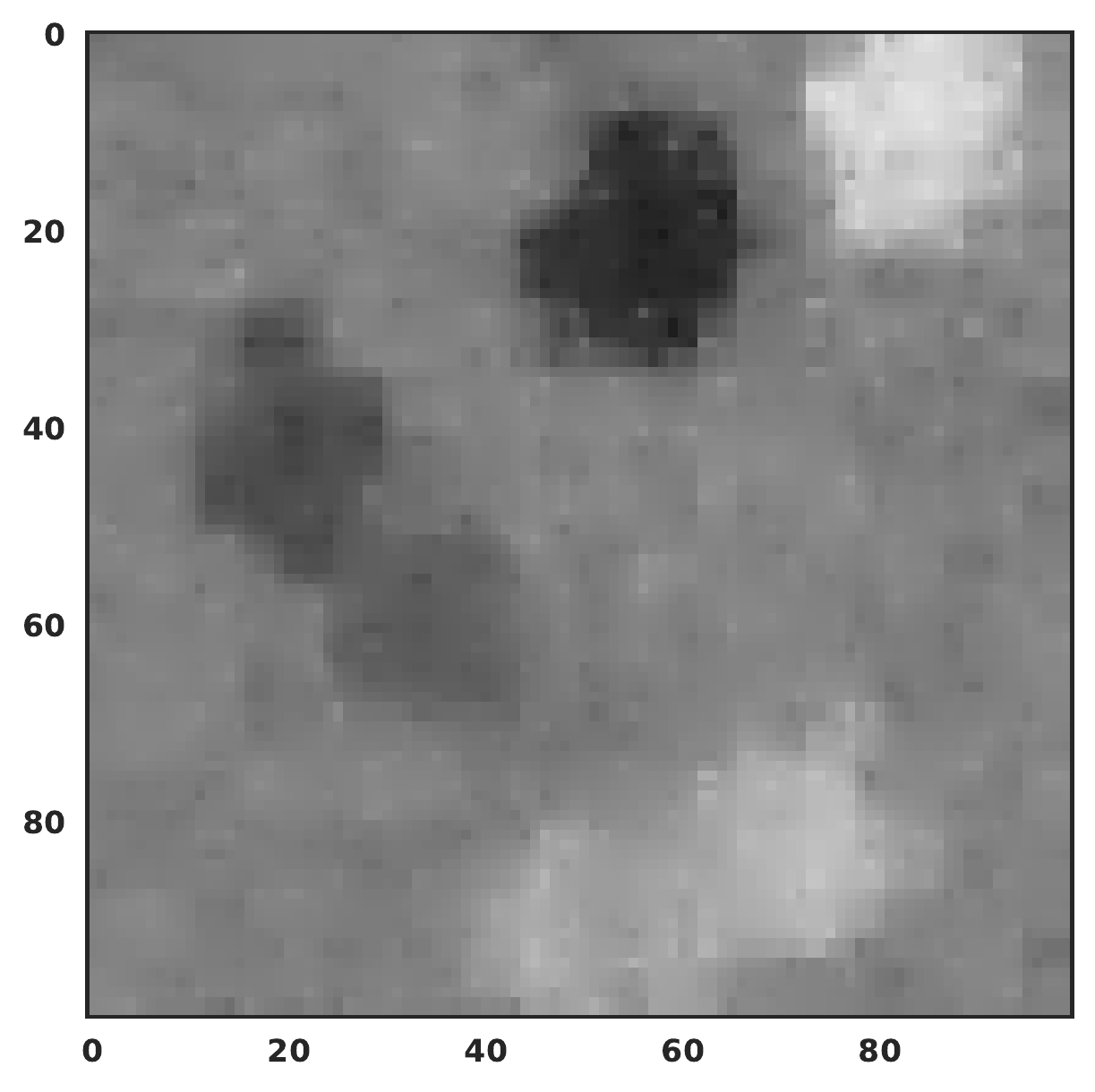}
\caption{CRISP, N = 2000}
\end{subfigure}
\begin{subfigure}[t]{0.32\textwidth}
\includegraphics[width=\textwidth]{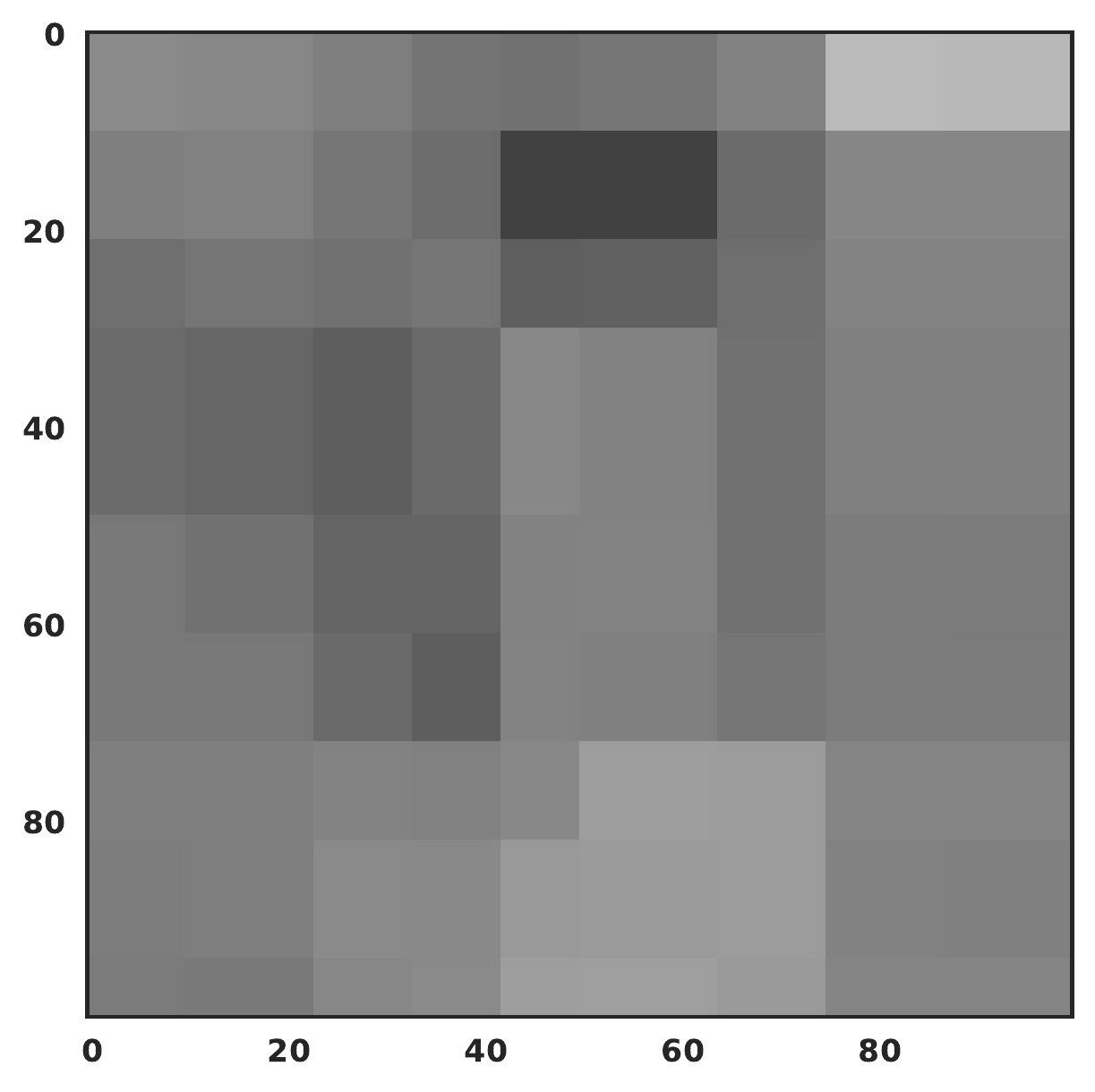}
\caption{GapCRISP, N = 100}
\end{subfigure}
\begin{subfigure}[t]{0.32\textwidth}
\includegraphics[width=\textwidth]{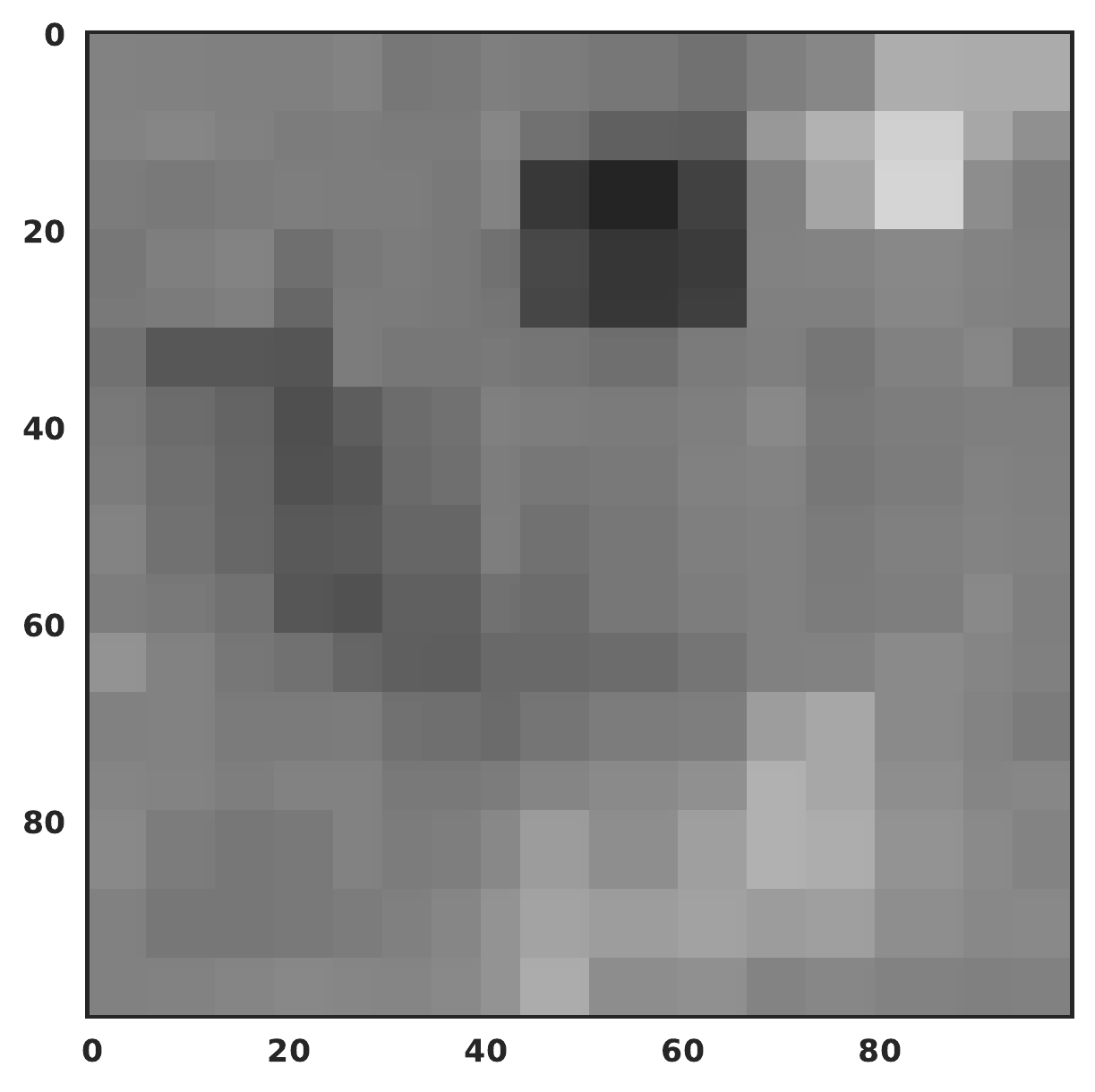}
\caption{GapCRISP, N = 500}
\end{subfigure}
\begin{subfigure}[t]{0.32\textwidth}
\includegraphics[width=\textwidth]{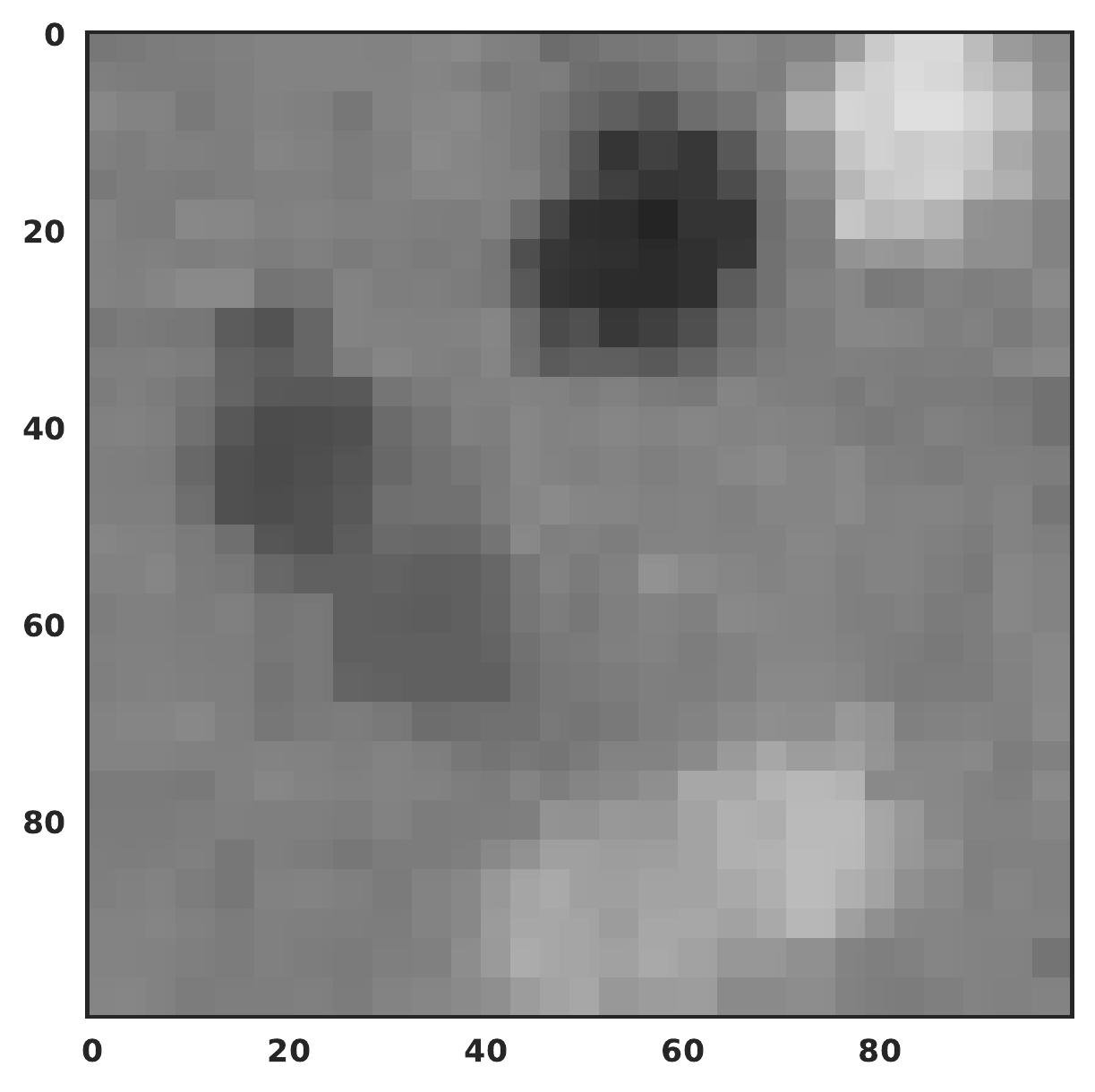}
\caption{GapCRISP, N = 2000}
\end{subfigure}
\begin{subfigure}[t]{0.32\textwidth}
\includegraphics[width=\textwidth]{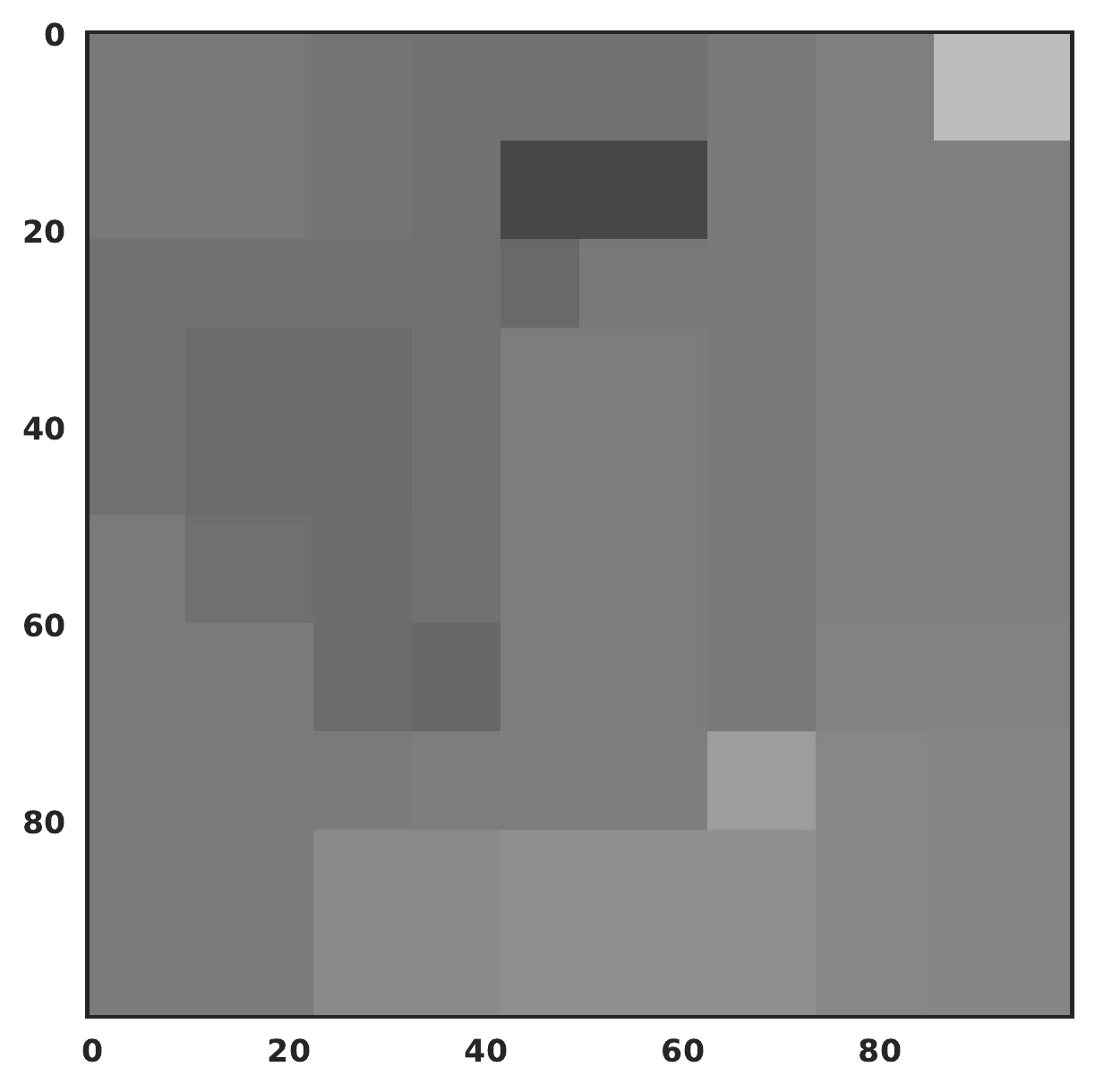}
\caption{GapTV, N = 100}
\end{subfigure}
\begin{subfigure}[t]{0.32\textwidth}
\includegraphics[width=\textwidth]{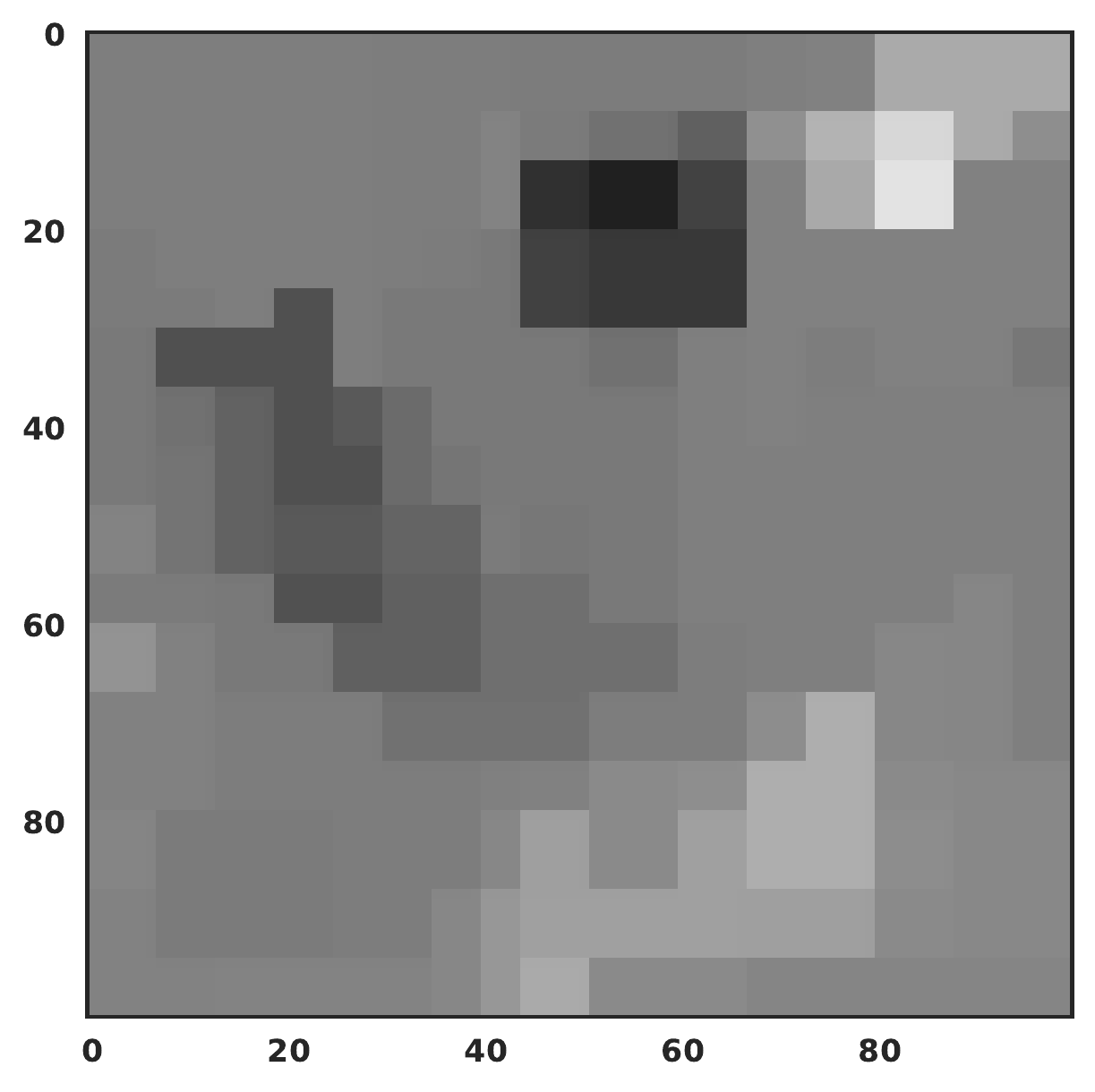}
\caption{GapTV, N = 500}
\end{subfigure}
\begin{subfigure}[t]{0.32\textwidth}
\includegraphics[width=\textwidth]{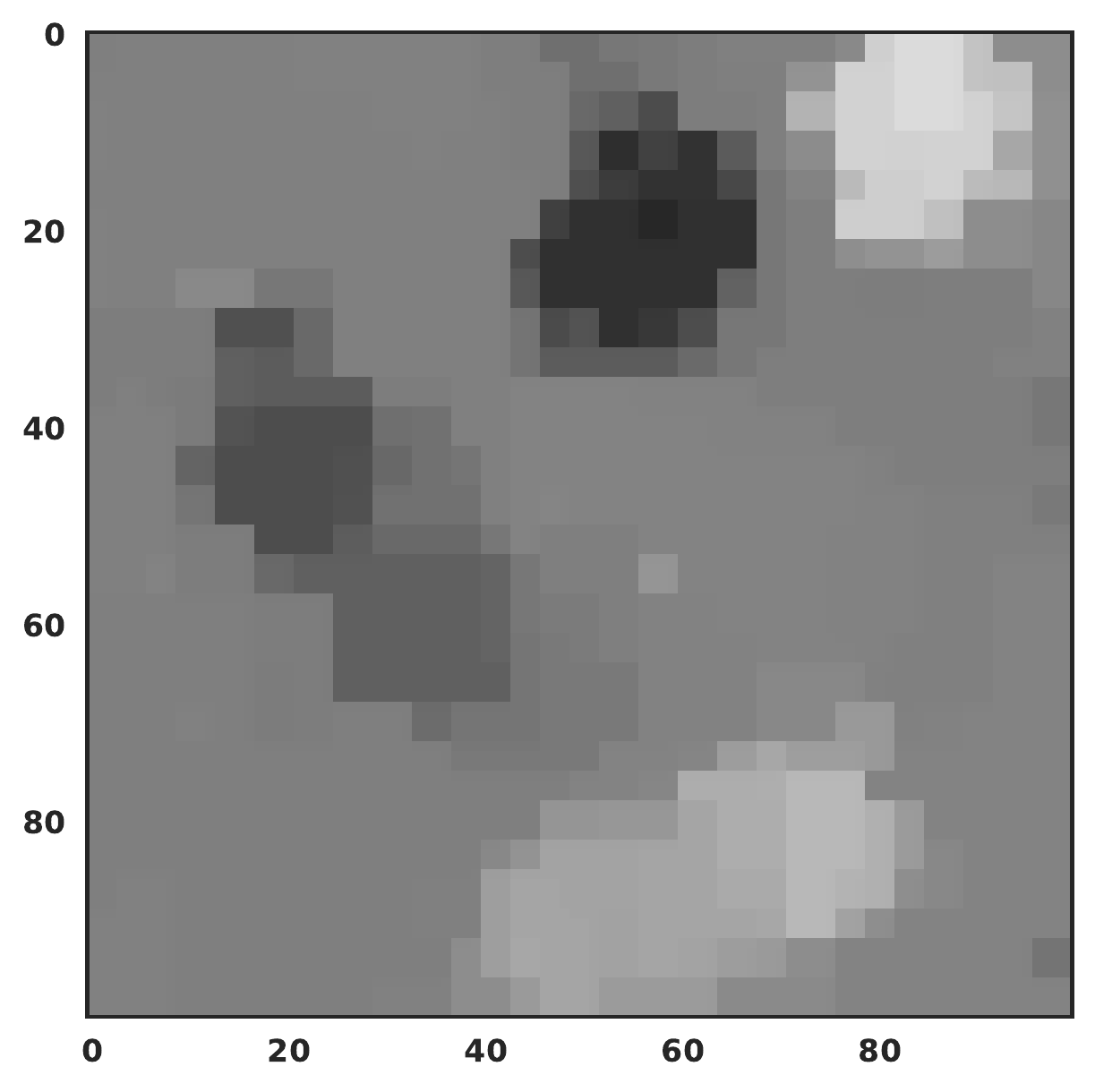}
\caption{GapTV, N = 2000}
\end{subfigure}
\caption{\label{fig:synthetic_results_qualitative} Qualitative examples of the four benchmark methods as the sample size increases. }
\end{figure*}

Finally, Figure \ref{fig:synthetic_results_qualitative} shows qualitative results for the four smoothing methods as the sample size grows from 100 to 2000. CART (Panels A-C) tends to over-smooth, leading to very sharp partitions that are too coarse grained to produce accurate results even as the sample size grows large. On the other hand, CRISP (Panels D-F) under-smooths by creating very blurry images. The gap-based version of CRISP (Panels G-I) alleviates this in the low-sample cases, but tying across entire rows and columns causes the image to blur as the data increases. The GapTV method (Panels J-L) achieves a reasonable balance here by producing large blocks in the low-sample setting and progressively refining the blocks as the sample size increases, without substantially compromising the sharpness of the overall image.

\begin{figure*}[!htb]
\centering
\begin{subfigure}[t]{0.32\textwidth}
\includegraphics[width=\textwidth]{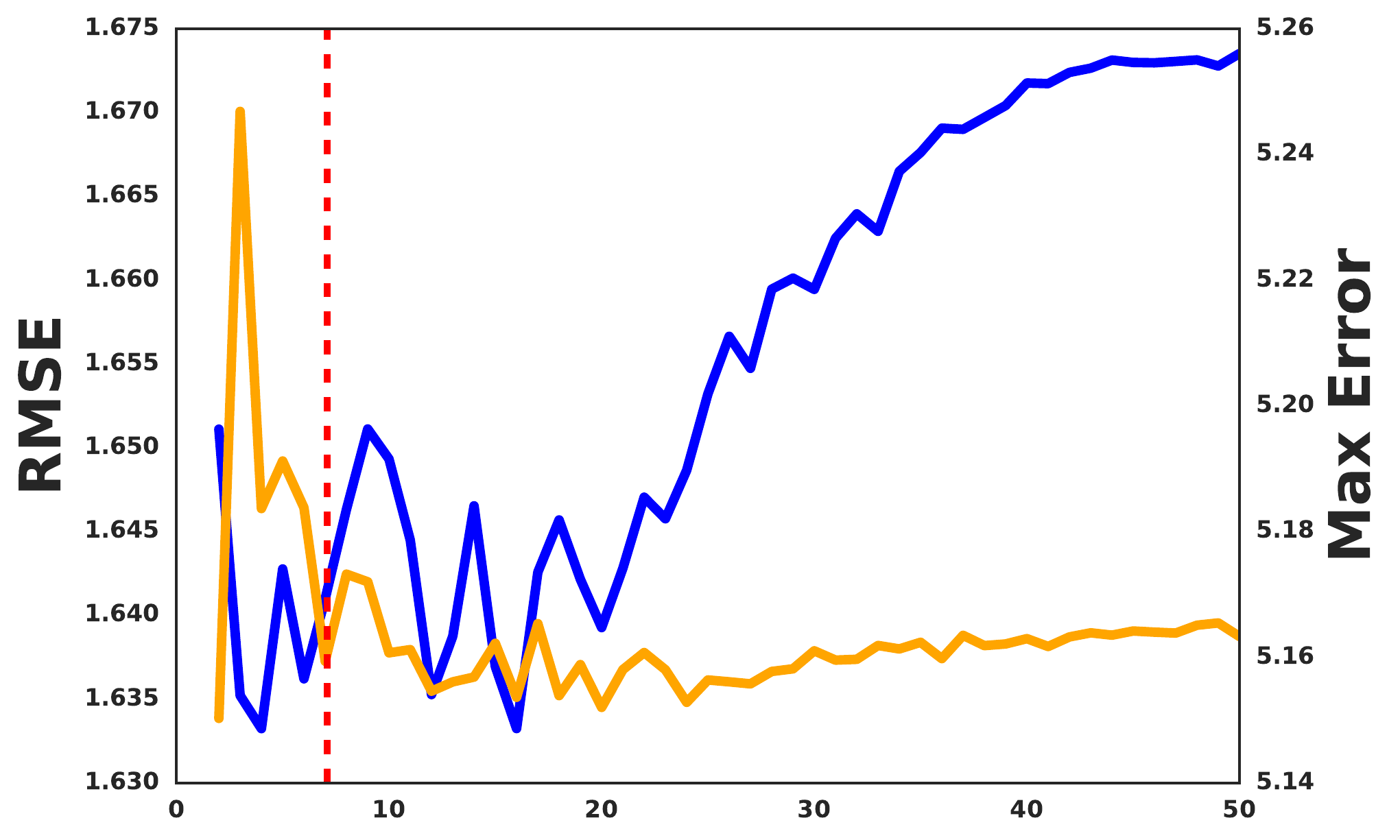}
\caption{N = 100}
\end{subfigure}
\begin{subfigure}[t]{0.32\textwidth}
\includegraphics[width=\textwidth]{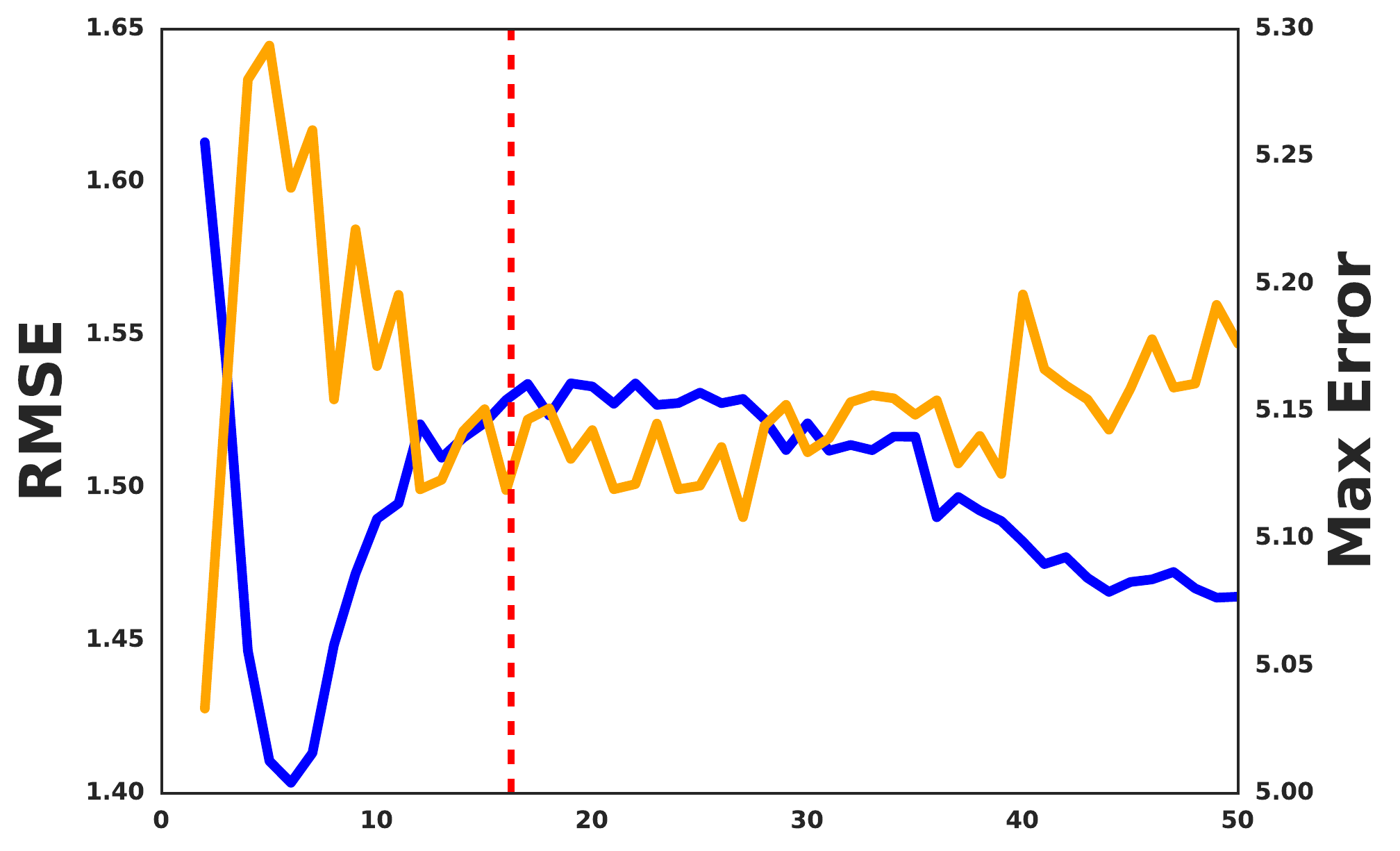}
\caption{N = 500}
\end{subfigure}
\begin{subfigure}[t]{0.32\textwidth}
\includegraphics[width=\textwidth]{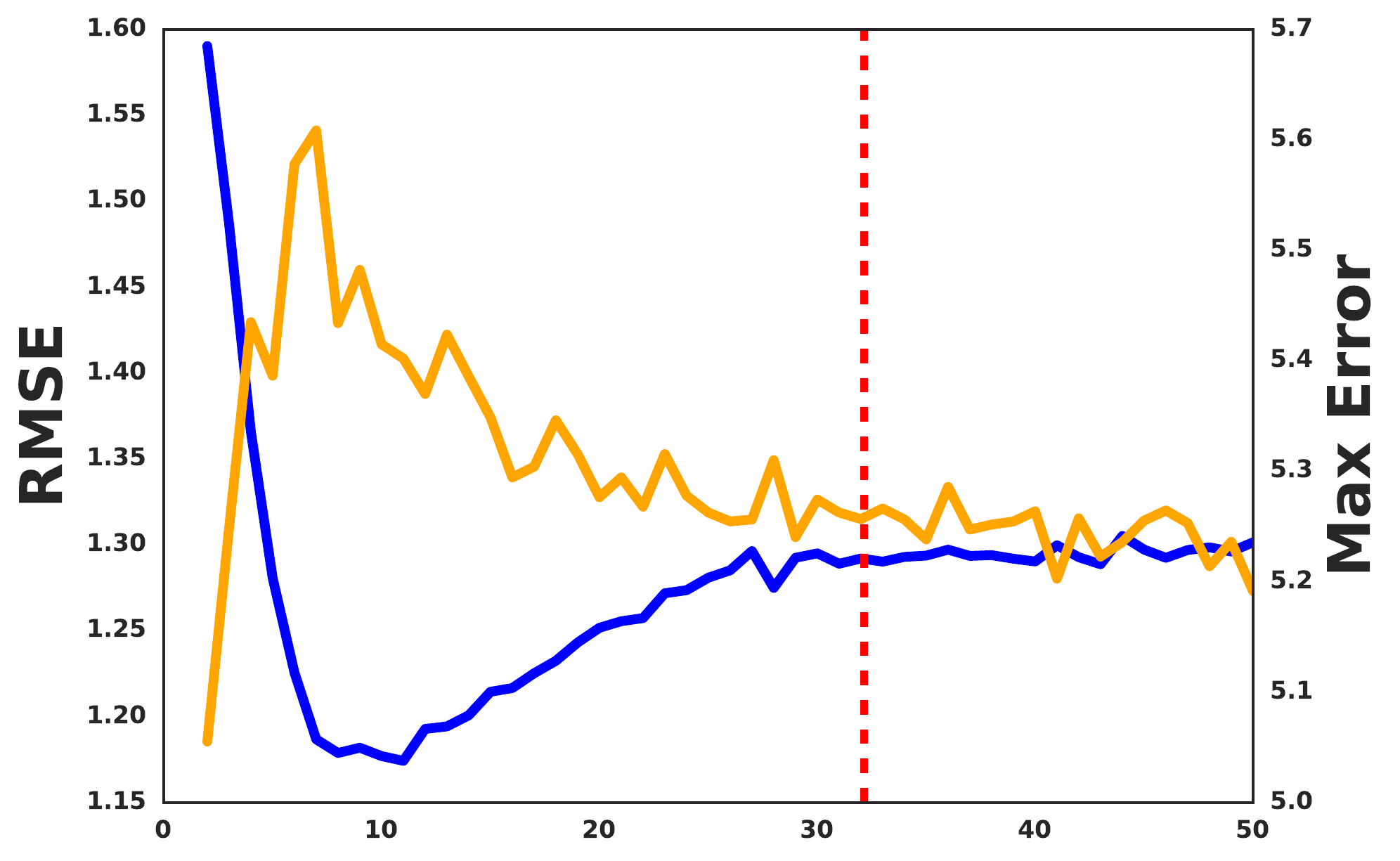}
\caption{N = 2000}
\end{subfigure}
\caption{\label{fig:gap_rmse_maxerr} RMSE (blue) and maximum error (orange) for the GapTV method for different sizes of the grid ($q^2$) for three different sample sizes; the dashed vertical red line indicates the value of $q$ chosen by the gap statistic. The results demonstrate that the gap statistic chooses models which provide a balance between average and worst-case error.\vspace{-0.2in}}
\end{figure*}

\subsection{Gap Statistic Evaluation}
\label{subsec:experiments:gap}

In order to understand the effect of the gap statistic, we conducted a series of synthetic benchmark experiments.  For each GapTV trial and sample size in the experiment from Section \ref{subsec:experiments:synthetic}, we exhaustively solved the graph TV problem for all possible values of $q$ in the range $[2,50]$. Figure \ref{fig:gap_rmse_maxerr} shows how the choice of $q$ impacts the average RMSE and maximum point error for three different sample sizes; the dotted vertical red line denotes the value selected by the gap statistic. As expected, when the sample size is small, the gap statistic selects much smaller values; as the sample size grows, the gap statistic selects progressively larger $q$ values. This enables the model to smooth over increasingly finer-grained resolutions.

Perhaps counter-intuitively, the gap statistic is \textit{not} choosing the $q$ value which will simply minimize RMSE. As the middle panel shows, the gap statistic may actually choose one of the worst possible $q$ values from this perspective. Instead, the resulting model is identifying a good trade-off between average accuracy (RMSE) and worst-case accuracy (max error). In small-sample scenarios like Figure \ref{fig:gap_rmse_maxerr}a, RMSE is not substantially impacted by having a very coarse-grained $q$. Thus this trade-off helps prevent over-smoothing in the small sample regime-- a problem observed by \cite{petersen:etal:2016} when using TV with a large $q$. As the data grows (Figure \ref{fig:gap_rmse_maxerr}b), both overly-fine and overly-coarse grids may have problems, with the latter now creating the potential for the TV method to under-smooth similarly to how CRISP performed in the synthetic benchmarks. Once sample sizes become relatively large (Figure \ref{fig:gap_rmse_maxerr}c), making the grid very fine-grained poses less risk of under-smoothing. The gap statistic here prevents $q$ from being chosen too low, which would create a much higher variance estimation.

\subsection{Austin and Chicago Crime Data}
\label{subsec:crime}

\begin{figure*}[t]
\centering
\begin{subfigure}[t]{0.45\textwidth}
\includegraphics[width=\textwidth]{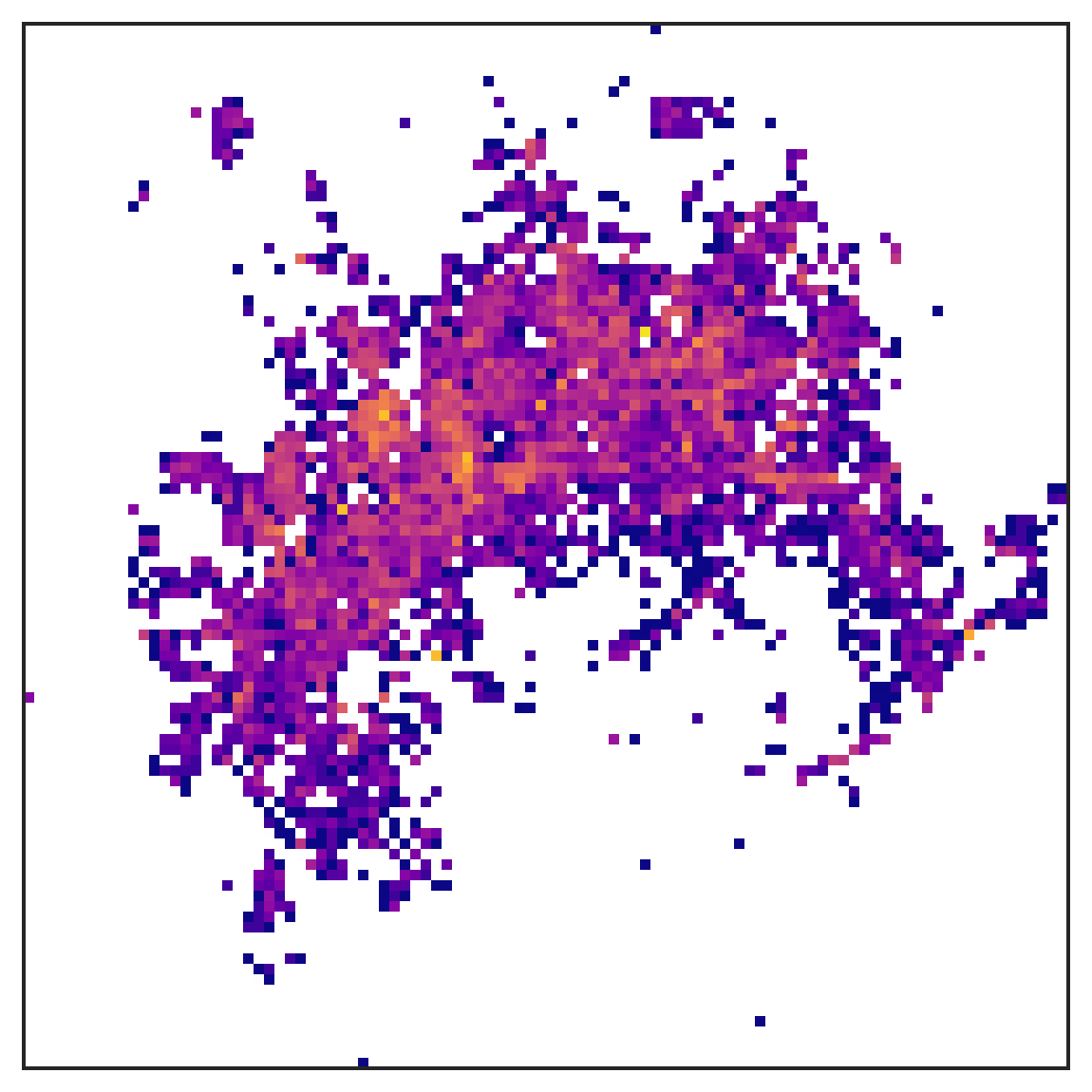}
\caption{Raw}
\end{subfigure}
\begin{subfigure}[t]{0.45\textwidth}
\includegraphics[width=\textwidth]{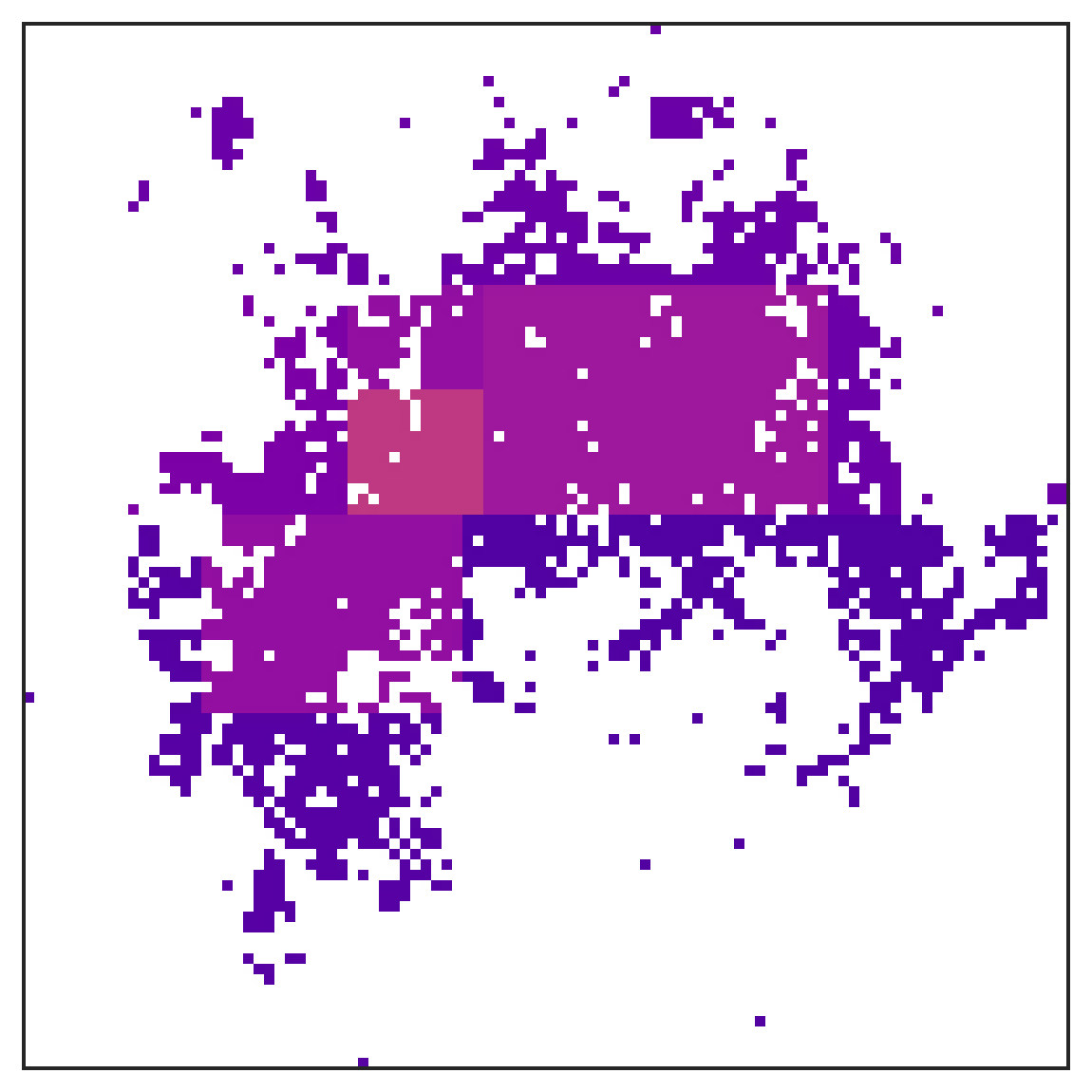}
\caption{CART}
\end{subfigure}
\begin{subfigure}[t]{0.45\textwidth}
\includegraphics[width=\textwidth]{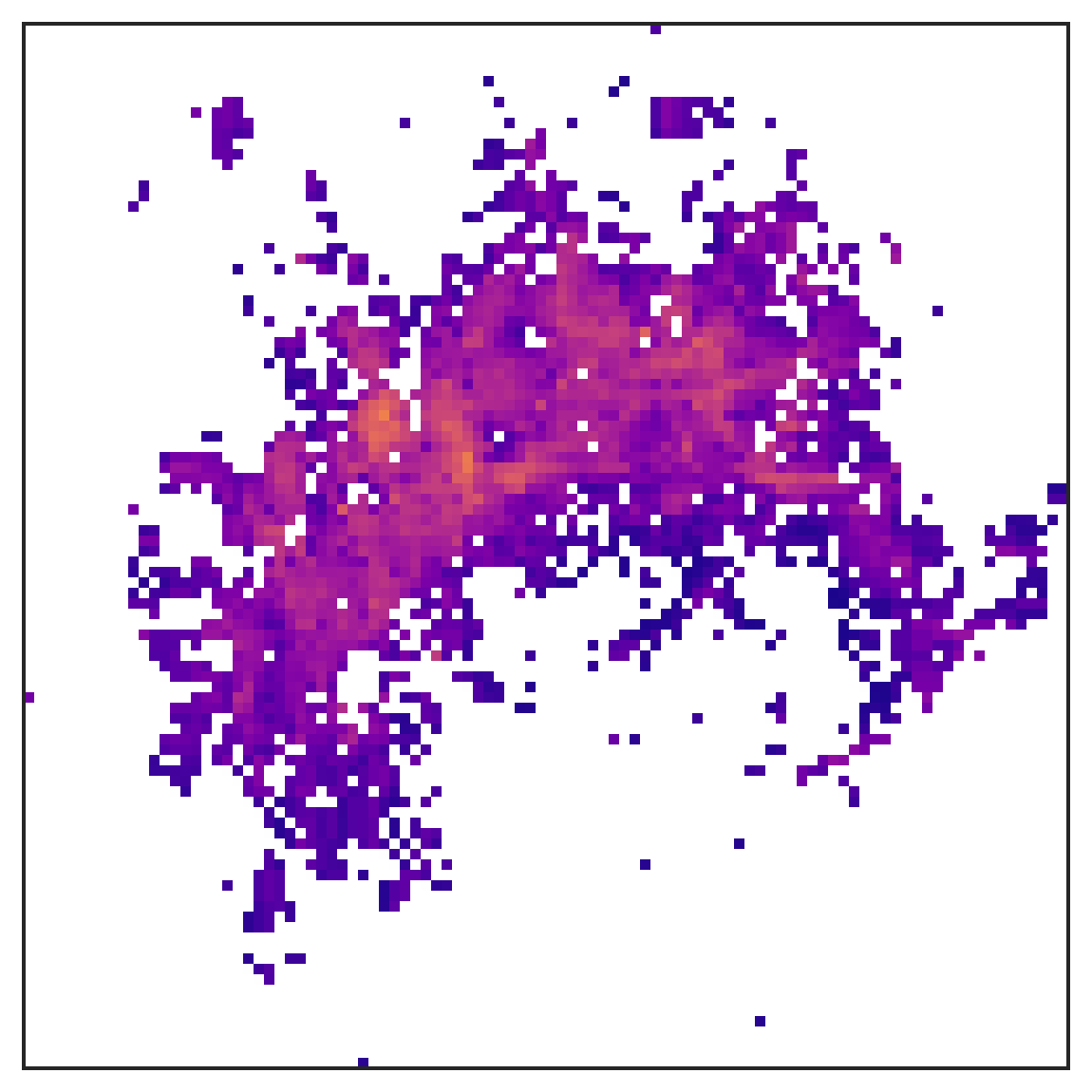}
\caption{CRISP}
\end{subfigure}
\begin{subfigure}[t]{0.45\textwidth}
\includegraphics[width=\textwidth]{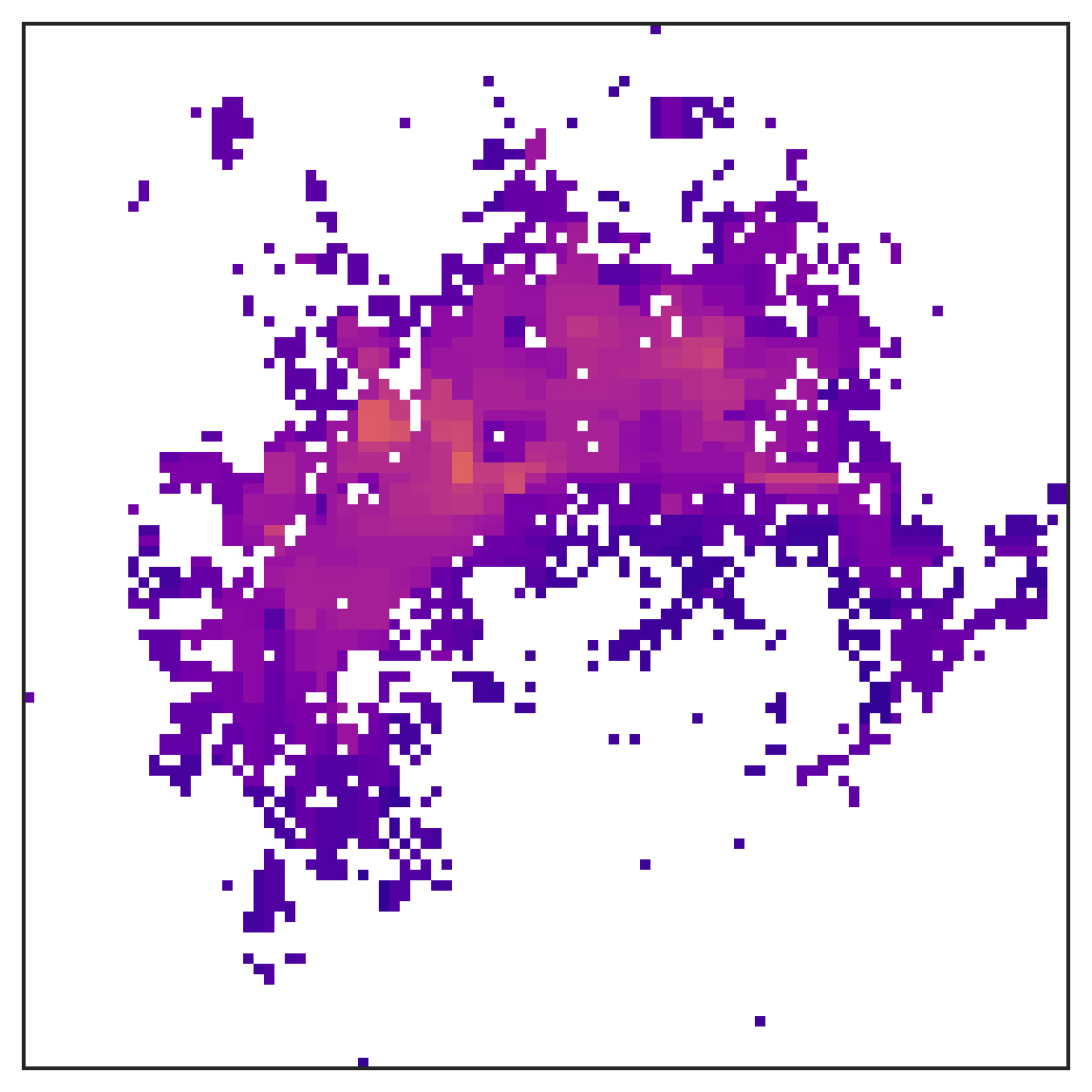}
\caption{GapTV}
\end{subfigure}
\caption{\label{fig:crime_results} Areal data results for the Austin crime data. The maps show the raw fine-grained results (Panel A) and the results of the three main methods. Qualitatively, CART (Panel B) over-smooths and creates too few regions in the city; CRISP (Panel C) under-smooths, creating too many regions; and GapTV (Panel D) provides a good balance that yields interpretable sections.}
\end{figure*}

As a final case study, we applied all four methods to a dataset of publicly-available crime report counts\footnote{\url{https://www.data.gov/open-gov/}} in Austin, Texas in 2014 and Chicago, Illinois in 2015. To preprocess the data, we binned all observations into a fine-grained $100 \times 100$ grid based on latitude and longitude, then took the log of the total counts in each cell. Points with zero observed crimes were omitted from the dataset as it is unclear whether they represented the absence of crime or a location outside the boundary of the local police department. Figure \ref{fig:crime_results} (Panel A) shows the raw data for Austin; the matching figure for Chicago is available in the appendix.

Each of the four methods considered in the previous sections were tested. The gap methods used $q$ values in the range $[2,100]$ and the CRISP method had $q=100$. To evaluate the methods, we ran a 20-fold cross-validation to measure RMSE and calculated plateaus with a fully-connected grid (i.e., as if all pixels were connected) which we then projected back to the real data for every non-missing point. Figure \ref{fig:crime_results} shows the qualitative results for CART (Panel B), CRISP (Panel C), and GapTV (Panel D); due to space considerations, GapCRISP is omitted as it adds little insight. The CART model clearly over-smooths by dividing the entire city into huge blocks of constant plateaus; conversely, CRISP under-smooths and creates too many regions. The GapTV method finds an appealing visual balance, creating flexible plateaus that partition the city well. These results are confirmed quantitatively in Table \ref{tab:crime_results}, where GapTV outperforms the three other methods in terms of AIC.

\begin{small}
\begin{table}
\centering
\begin{tabular}{|l|lll|}
\hline
\multicolumn{1}{|c}{} & \multicolumn{3}{|c|}{Austin Crime Data} \\ \hline
 & RMSE & Plateaus & AIC \\ \hline
CART & 1.0522 & 10.4000 & 11139.2911 \\
CRISP & 0.9420 & 4699.1500 & 18326.3333 \\
GapCRISP & 0.9633 & 1361.7500 & 12064.2507 \\
GapTV & 0.9743 & 384.3500 & 10327.5860 \\
\hline
\multicolumn{4}{c}{}\\
\hline
\multicolumn{1}{|c}{} & \multicolumn{3}{|c|}{Chicago Crime Data} \\ \hline
 & RMSE & Plateaus & AIC \\ \hline
CART & 1.0460 & 9.2500 & 43804.6942 \\
CRISP & 0.8450 & 9330.6000 & 47245.5734 \\
GapCRISP & 0.8476 & 8278.9000 & 45314.7106 \\
GapTV & 0.8581 & 2270.1500 & 34016.5952 \\
\hline
\end{tabular}
\caption{\label{tab:crime_results} Quantitative results for the four methods on crime data for Austin and Chicago. The GapTV method achieves the best trade-off between accuracy and the number of constant regions, as measured by AIC.}
\end{table}
\end{small}

%% file: conclusion.tex
\vspace{-0.1in}
\section{Conclusion}
\label{sec:conclusion}

This paper presented GapTV, a new method for interpretable low-dimensional regression. Through a novel use of the gap statistic, our model divides the covariate space into a finite-sized grid in a data-adaptive manner. We then use a fast TV denoising algorithm to smooth over the cells, creating plateaus of constant value. On a series of synthetic benchmarks, we demonstrated that our method produces superior results compared to a baseline CART model and the current state of the art (CRISP). Finally, we provided additional evaluation through a real-world case study on crime rates in Austin and Chicago, showing that GapTV discovers much more interpretable and meaningful spatial plateaus. Overall, we believe the speed, accuracy, interpretability, and fully auto-tuned nature of GapTV makes it a strong candidate for low-dimensional regression.

%% file: app_chicago.tex
\section{Chicago Results}
\label{sec:app:chicago}
Below are the results for the three main methods applied to the Chicago data.
\begin{figure*}[!h]
\centering
\begin{subfigure}[t]{0.45\textwidth}
\includegraphics[width=\textwidth]{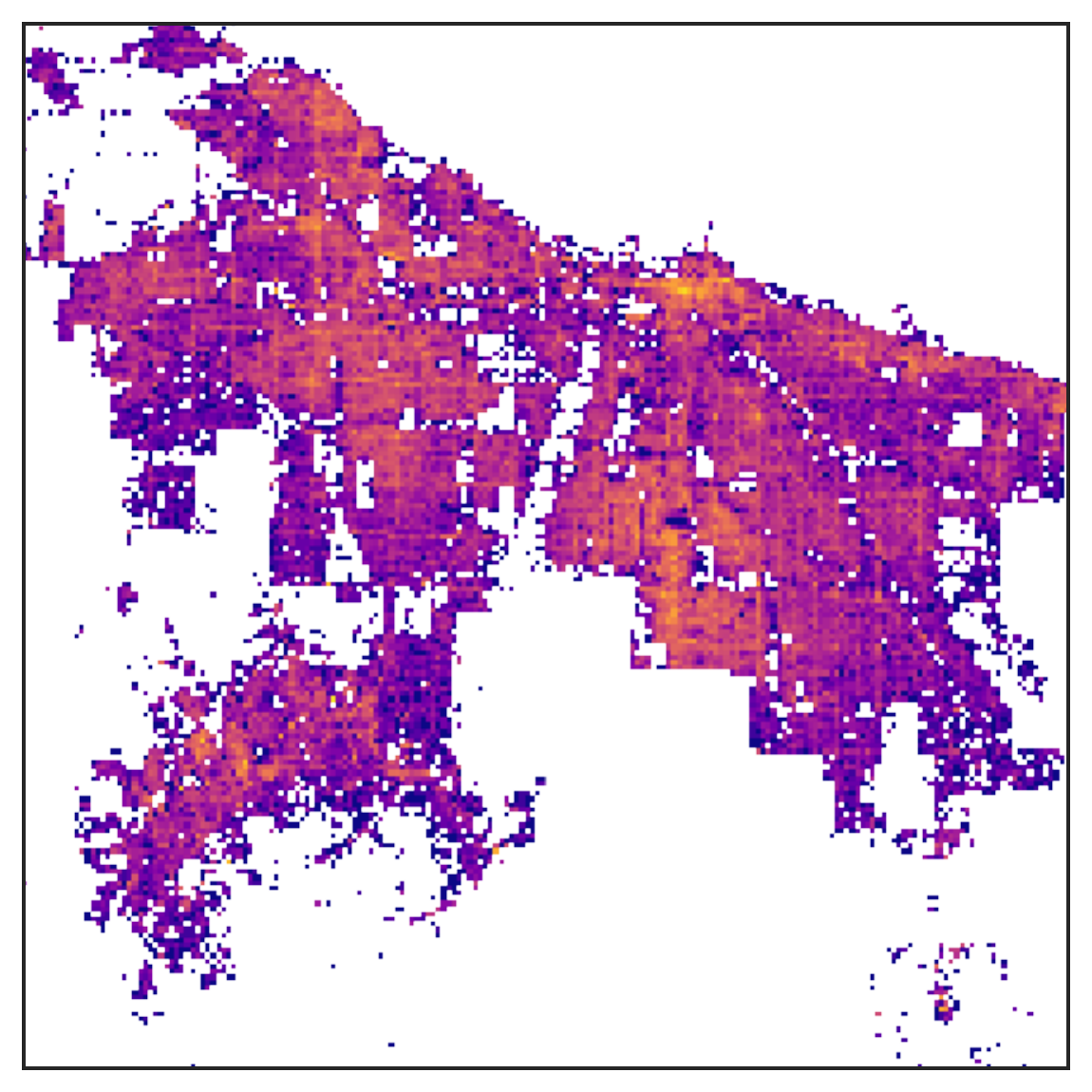}
\caption{Raw}
\end{subfigure}
\begin{subfigure}[t]{0.45\textwidth}
\includegraphics[width=\textwidth]{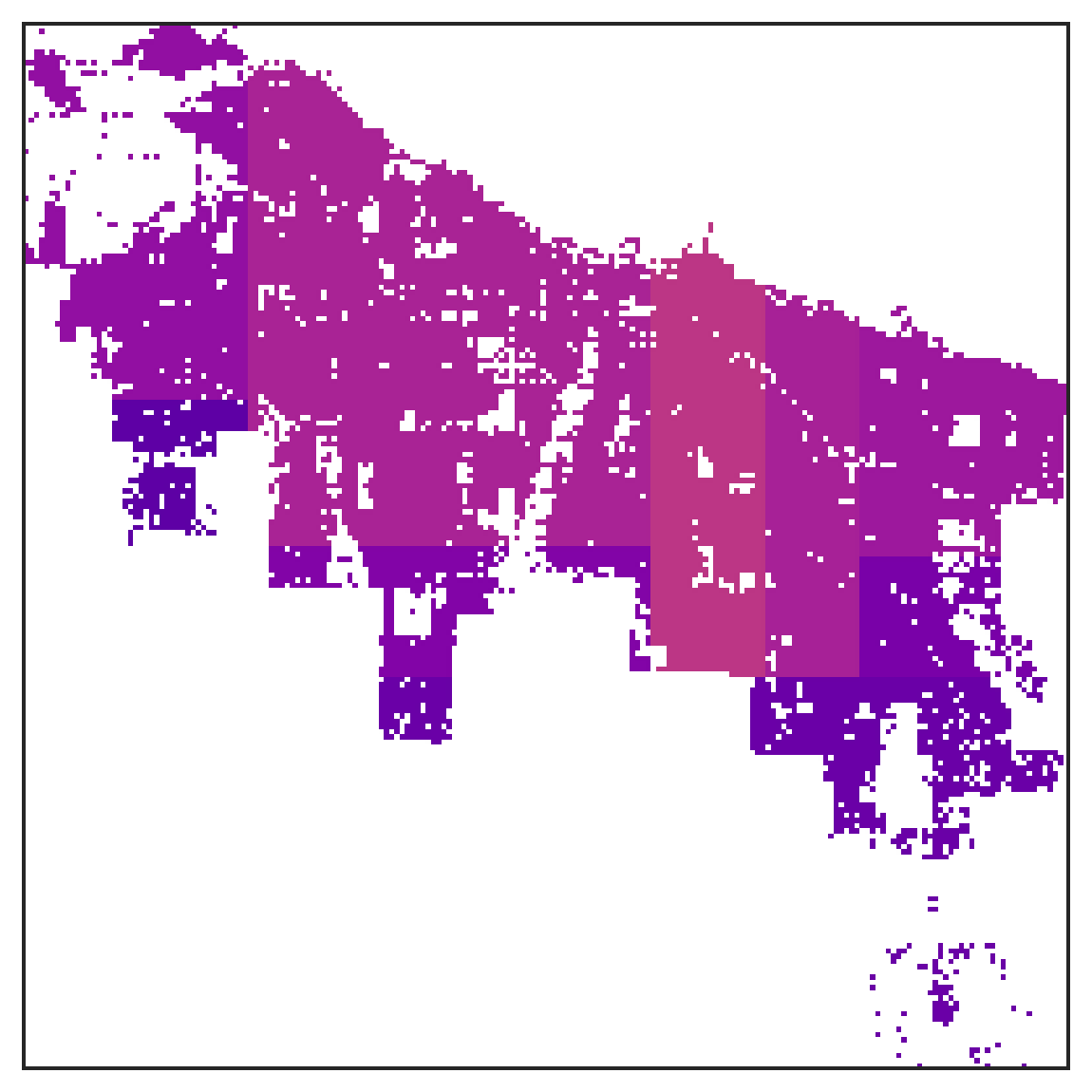}
\caption{CART}
\end{subfigure}
\begin{subfigure}[t]{0.45\textwidth}
\includegraphics[width=\textwidth]{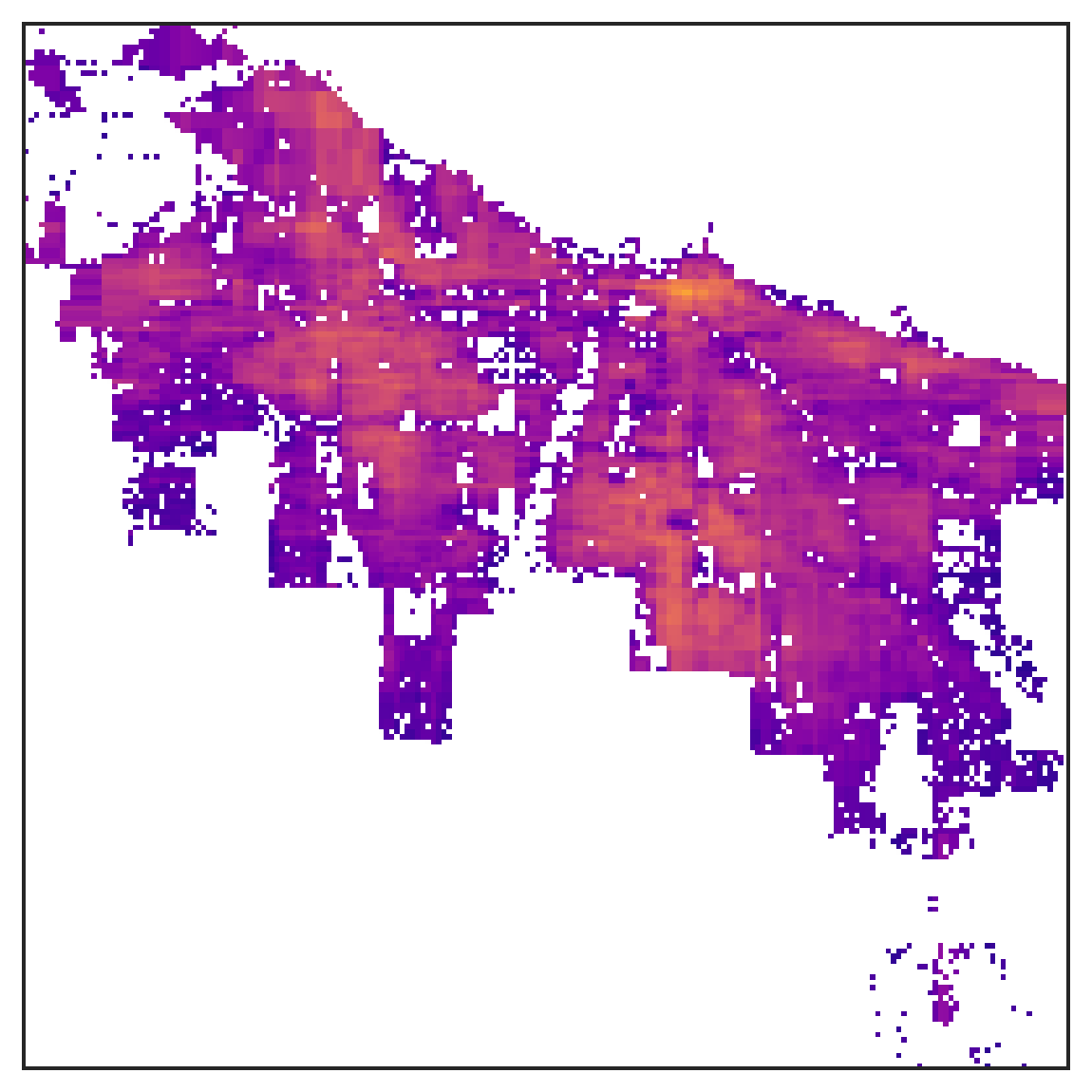}
\caption{CRISP}
\end{subfigure}
\begin{subfigure}[t]{0.45\textwidth}
\includegraphics[width=\textwidth]{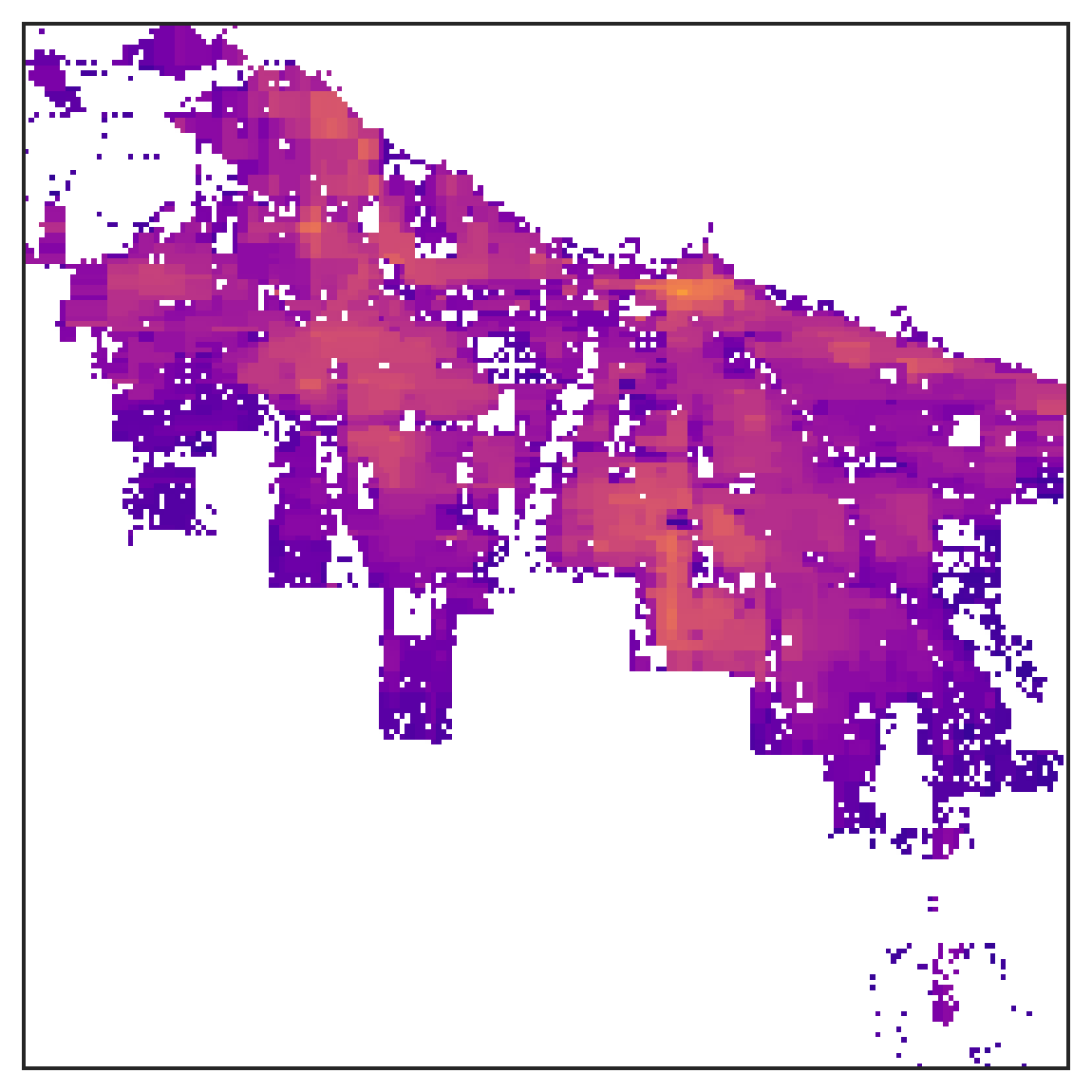}
\caption{GapTV}
\end{subfigure}
\caption{\label{fig:crime_results_chicago} Areal data results for the Chicago crime data. The maps show the raw fine-grained results (Panel A) and the results of the three main methods. Qualitatively, CART (Panel B) over-smooths and creates too few regions in the city; CRISP (Panel C) under-smooths, creating too many regions; and GapTV (Panel D) provides a good balance that yields interpretable sections.}
\end{figure*}